%% file: main.tex
\documentclass[10pt,twocolumn,letterpaper]{article}

\usepackage[pagenumbers]{cvpr}
\input{preamble}

\definecolor{cvprblue}{rgb}{0.21,0.49,0.74}
\usepackage[pagebackref,breaklinks,colorlinks,allcolors=cvprblue,hypertexnames=false]{hyperref}
\hypersetup{
  pdftitle={Totem: Object-Centric 3D Memory for World-Consistent Video Generation},
  pdfauthor={Zichen Wan, Tianyu Huang, Fan Li, Kai Wang, Hongzhi Zhang, Wangmeng Zuo}
}

\title{Totem: Object-Centric 3D Memory for World-Consistent Video Generation}
\author{%
  Zichen Wan\textsuperscript{1}\quad
  Tianyu Huang\textsuperscript{2}\quad
  Fan Li\textsuperscript{2}\\
  Kai Wang\textsuperscript{1}\quad
  Hongzhi Zhang\textsuperscript{1}\quad
  Wangmeng Zuo\textsuperscript{1}\\
  {\normalsize\textsuperscript{1}Harbin Institute of Technology\quad
  \textsuperscript{2}Huawei Technologies Ltd.}
}

\begin{document}
\makeatletter
\twocolumn[{%
  \@maketitle
  \centering
  \includegraphics[width=\textwidth]{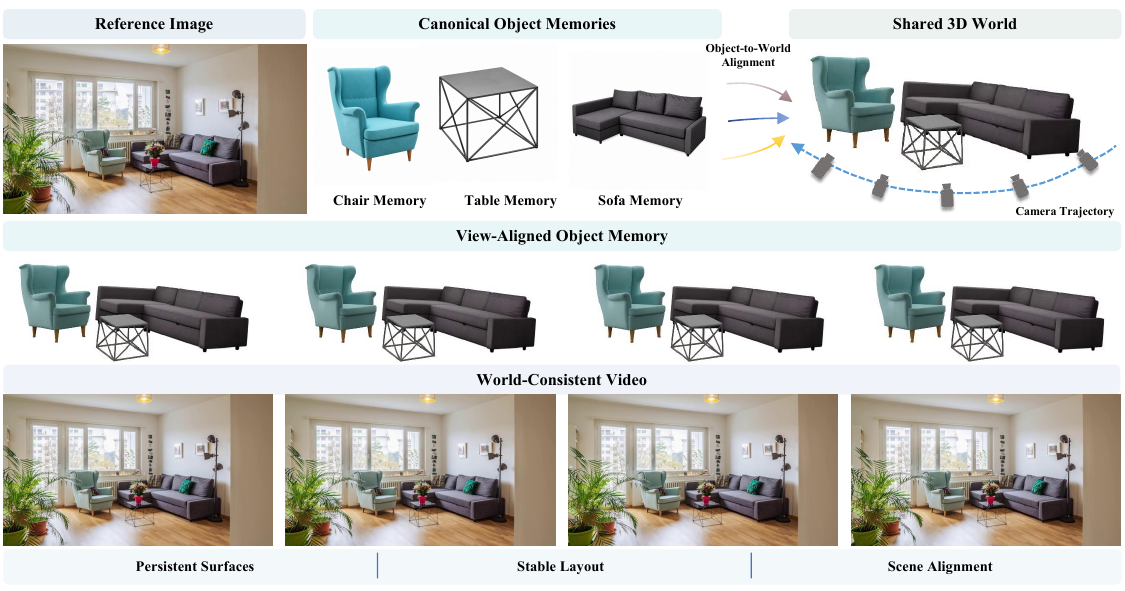}
  \def\@captype{figure}\caption{Totem is an object-centric 3D memory framework for world-consistent video generation.
  Up: Totem builds canonical memories for selected objects from a reference image and establishes their object-to-world transforms in a shared 3D world. Bottom: Given custom camera trajectories, Totem renders view-aligned object memories to generate world-consistent videos with persistent single-object surfaces, stable inter-object layouts, and consistent object-scene alignment.}\label{fig:teaser}
  \vspace{\dbltextfloatsep}
}]
\makeatother

\input{sec/abstract}
\input{sec/introduction}
\input{sec/related_work}
\input{sec/method}
\input{sec/experiments}
\input{sec/applications}
\input{sec/conclusion}

{\small
\bibliographystyle{ieeenat_fullname}
\bibliography{references}
}

\clearpage
\setcounter{page}{1}
\appendix
\maketitlesupplementary
\raggedbottom
\input{sec/supplementary_material}

\end{document}

%% file: preamble.tex
\usepackage{graphicx}
\usepackage{amsmath}
\usepackage{amssymb}
\usepackage{booktabs}
\usepackage{array}
\usepackage{colortbl}
\usepackage{tabularx}
\usepackage{algorithm}
\usepackage{algorithmic}

\definecolor{TableGroup}{gray}{0.94}

\newcolumntype{Y}{>{\centering\arraybackslash}X}
\newcolumntype{H}[1]{>{\centering\arraybackslash
  \advance\leftskip by -#1\relax\advance\rightskip by #1\relax}X}
\newcolumntype{P}{>{\hsize=.92\hsize\centering\arraybackslash}X}
\newcolumntype{Q}{>{\hsize=1.08\hsize\centering\arraybackslash}X}
\newlength{\tablecolgap}
\newsavebox{\tablemeasurebox}
\newcommand{\tableendpad}{}
\newcolumntype{S}[1]{@{\hspace{\dimexpr\tablecolgap-#1\relax}}}
\newcolumntype{G}[2]{@{%
  \hspace{\dimexpr(\tablecolgap-0.4pt)/2-#1\relax}%
  \vrule width 0.4pt%
  \hspace{\dimexpr(\tablecolgap-0.4pt)/2-#2\relax}%
}}
\NewDocumentEnvironment{equalspacedtabular}{o m m +b}{%
  \setlength{\tabcolsep}{0pt}%
  \setlength{\tablecolgap}{0pt}%
  \IfNoValueTF{#1}{%
    \renewcommand{\tableendpad}{}%
  }{%
    \renewcommand{\tableendpad}{\hspace{\dimexpr(\tablecolgap-0.4pt)/2-#1\relax}}%
  }%
  \sbox{\tablemeasurebox}{\begin{tabular}{@{}#3@{\tableendpad}}#4\end{tabular}}%
  \IfNoValueTF{#1}{%
    \setlength{\tablecolgap}{\dimexpr(\linewidth-\wd\tablemeasurebox)/#2\relax}%
  }{%
    \setlength{\tablecolgap}{\dimexpr(2\linewidth-2\wd\tablemeasurebox)/\numexpr2*#2+1\relax\relax}%
  }%
  \begin{tabular}{@{}#3@{\tableendpad}}#4\end{tabular}%
}{}
\newcommand{\tablehead}[1]{\begin{tabular}[c]{@{}c@{}}#1\end{tabular}}
\newcommand{\paperfigurelabel}[1]{{\normalfont\normalsize\color{black}#1}}

\newlength{\floatcontentsep}
\newcommand{\papertablestyle}{%
  \renewcommand{\arraystretch}{1.05}%
  \setlength{\tabcolsep}{0.8pt}%
}
\AtBeginEnvironment{table*}{\papertablestyle}

%% file: sec/abstract.tex
\begin{abstract}
    Generating coherent videos grounded in a unified underlying world remains a major challenge, particularly under large viewpoint changes, where existing methods often struggle to maintain stable representations of individual objects. To this end, we introduce \textbf{Totem}, an object-centric 3D memory framework for world-consistent video generation. Totem first represents each selected object as a persistent 3D memory aligned to a shared world frame, assigning it a dedicated state. Based on these memories, we design a dual-stream 3D conditioning architecture that combines object memories with scene geometry for target-view synthesis. Moreover, these object memories constrain the model’s attention with 3D structural information, thereby preserving
    object surfaces and relative spatial layouts. Extensive experiments on world-generation and object-centric benchmarks demonstrate that Totem generates videos depicting a more object-consistent 3D world under challenging camera trajectories, while supporting versatile applications.
    \end{abstract}

%% file: sec/introduction.tex
\section{Introduction}

Video generation is becoming a fundamental capability for a wide range of visual applications, from content creation to interactive and immersive experiences. Recent image-to-video models provide a flexible way to generate realistic videos from a single reference image~\citep{blattmann2023stable,hacohen2024ltxvideo,yang2024cogvideox,kong2024hunyuanvideo,wan2025wan}. To support a user-defined camera trajectory, the generation process is further conditioned on camera information such as poses~\citep{wang2024motionctrl} and ray representations~\citep{he2024cameractrl,bahmani2024ac3d}.
Although these methods improve viewpoint controllability, they mainly learn it as image-space motion, making it difficult to maintain a consistent world under complex viewpoint changes.

Recently, increasing efforts have explored world-consistent video generation through explicit 3D memory. These methods attempt to guide video generation by rendering target-view observations from scene-level point clouds~\citep{yu2024viewcrafter,cao2025uni3c,ren2025gen3c,huang2025voyager,zhao2025spatia,wang2026anchorweave,zhang2026worldstereo}, but the resulting memories usually cover only the geometry visible in the input views. When the camera moves beyond the observed regions, the rendered conditions provide limited support for maintaining stable objects in the generated world. 
As a result, existing methods still face several object-level challenges under complex scenarios with occlusions, object disappearance and reappearance, and strong scale variation. 1) \textbf{Object-Scene Alignment Drift.} Without coordination between object and scene, a target object may be wrongly hidden or misaligned with surrounding structures.
2) \textbf{Single-Object Surface Inconsistency.} Scene-level memory does not reliably associate generated pixels with the same physical object surfaces. Consequently, object appearance and geometry may become inconsistent.
3) \textbf{Inter-Object Layout Instability.} Spatial relations among individual objects are not explicitly modeled by existing conditions, leading to unstable layouts.

Addressing these challenges requires more than view-dependent scene guidance. What is needed is a persistent object-level representation that can carry object information across viewpoints while retaining its geometric context within the scene. This naturally motivates an object-centric 3D memory, which provides a stable spatial reference for individual objects within a shared 3D world. This allows object properties to remain associated with the same underlying entities, while helping preserve inter-object layouts and object-scene alignment.

In this work, we introduce Totem, an object-centric 3D memory framework for world-consistent video generation from a single image along user-specified camera trajectories. Starting from the reference image, each selected object is instantiated as a 3D memory, comprising a canonical representation with surface anchors and an object-to-world transform that places it in a shared world coordinate system. To translate these memories into view-aligned guidance for generation, Totem adopts a dual-stream conditioning design that couples instance-level appearance and geometry with scene-level spatial structure, providing a stronger 3D prior for world-consistent synthesis. Furthermore, the 3D structure stored in memory enables surface correspondence and spatial relation modeling. We therefore propose object-centric attention in canonical and world spaces to exploit this structural information during generation: canonical-space surface binding associates repeated observations with canonical surface anchors; world-space relation preservation incorporates relative object transforms into cross-object attention.

We extensively evaluate Totem on both world-level and object-level benchmarks, including WorldScore~\citep{duan2025worldscore}, ConsIDVid-Bench~\citep{wu2026considgen}, and an out-of-distribution benchmark. Totem achieves state-of-the-art performance in world-consistent generation. Dual-stream view guidance and object-centric attention further bring strong improvements in cross-view object appearance and geometry under challenging scenarios. In addition, Totem supports applications such as camera-controllable video generation, long-range video generation, and spatial video editing.

Our contributions are summarized as follows:
\begin{itemize}
\item We introduce Totem, an object-centric 3D memory framework for world-consistent video generation, which maintains persistent 3D object states.
\item We propose a memory-guided generation architecture that transforms object-centric 3D memories into view-aligned priors and object-aware attention signals, stabilizing object appearance, geometry, and spatial relations across generated views.  
\item Totem achieves competitive performance across different tasks, demonstrating its potential as a world-consistent video diffusion model.
\end{itemize}

%% file: sec/related_work.tex
\section{Related work}

\noindent\textbf{Feed-forward 3D reconstruction.} Recent feed-forward 3D reconstruction methods have shifted 3D perception from optimization-heavy pipelines toward direct prediction from sparse observations. They recover camera parameters, depth, pointmaps, tracks, or global 3D structure through a unified forward pass~\citep{wang2024dust3r,leroy2024mast3r,wang2025vggt,wang2025pi3,lin2025depthanything3,keetha2025mapanything}. These methods provide efficient geometric priors and scene-level structure for downstream vision tasks. Moving beyond holistic scene reconstruction, object-centric 3D methods further focus on modeling individual objects from images, capturing detailed geometry and appearance~\citep{hong2023lrm,xu2024instantmesh,chen2025sam3d,du2026och3r,xiang2024trellis,zhao2025hunyuan3d,huang2025spar3d}. While they provide object-level 3D representations for reconstruction and rendering, they are primarily designed as static object assets rather than temporal memory for video generation.

\noindent\textbf{Video generation models.}
Diffusion video models established the foundation for modern video generation by incorporating temporal modeling into the diffusion framework~\citep{ho2022videodiffusion,ho2022imagenvideo,singer2022makeavideo}. Building on this foundation, subsequent latent diffusion models (LDMs) improved generation efficiency and quality, while enabling conditional video synthesis~\citep{blattmann2023videoldm,guo2023animatediff,blattmann2023stable,chen2024videocrafter,zhang2023i2vgenxl}. More recently, large-scale video foundation models, often built on Transformer architectures, have further strengthened spatio-temporal modeling, visual fidelity, and motion realism through extensive training data~\citep{yang2024cogvideox,kong2024hunyuanvideo,hacohen2024ltxvideo,wan2025wan}. Overall, these models provide strong generative priors for realistic video synthesis and controllable motion.

\noindent\textbf{Spatial memory for video generation.} To improve spatial consistency, a growing body of work augments video diffusion pipelines with explicit spatial memory structures that maintain a 3D scene representation during generation. A representative strategy is to lift observed or generated RGB frames into scene point clouds using estimated depth, accumulate them into a persistent 3D cache, and render view-aligned conditions along the target camera trajectory~\citep{yu2024viewcrafter,cao2025uni3c,ren2025gen3c,huang2025voyager,zhao2025spatia,zhang2026worldstereo}. Beyond global scene caches, recent methods explore locally retrieved geometric memories from historical views~\citep{wang2026anchorweave}, while others construct persistent spatial memory directly in the diffusion latent space~\citep{wang2026mirage}. Such memory-based approaches have demonstrated clear improvements in multi-view coherence, as the accumulated 3D representation anchors newly generated frames to a shared geometric scaffold~\citep{wu2025videoworld}. Nevertheless, most existing memory designs organize 3D information at the scene level, without assigning persistent states to individual objects. Our work instead queries object-centric 3D memories to guide target-view synthesis, supporting long-range camera motion while preserving object-level consistency.

%% file: sec/method.tex
\section{Method}

Given a reference image \(I_0 \in \mathbb{R}^{3 \times H \times W}\), a text prompt \(y\), and a user-defined camera trajectory \(\mathcal{C}=\{C_t\}_{t=0}^{T-1}\), our goal is to generate a world-consistent video that preserves object-level consistency across changing viewpoints. However, three challenges remain: (1) the drift of object-scene alignment, (2) the inconsistency of object surfaces, and (3) the instability of inter-object layouts. To address them, we introduce Totem, an object-centric 3D memory framework. In this section, we first construct persistent 3D memories for selected objects, storing their renderable representations, canonical surface anchors, and object-to-world transforms (Sec.~\ref{sec:object-centric-memory}). Accordingly, we develop dual-stream view guidance by rendering object memories along the camera trajectory and complementing them with scene depth (Sec.~\ref{sec:view-guidance}). We further introduce object-centric attention to support fine-grained object modeling with 3D constraints (Sec.~\ref{sec:object-centric-attention}).

\begin{figure*}[t]
\centering
\includegraphics[width=\textwidth]{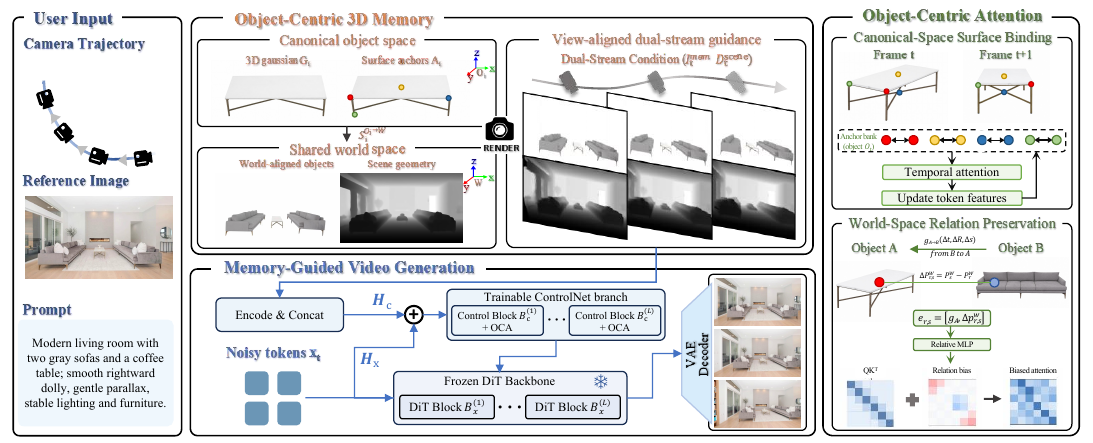}
\caption{Overview of Totem, arranged from left to right. Standard video-network details and routine injection paths for text, camera, image, and mask information are omitted for clarity. From the reference image and target camera trajectory on the left, Totem constructs canonical object memories, aligns them to a shared world frame, and renders view-aligned dual-stream conditions. In the center, the conditions are separately encoded and projected, fused into control tokens, and injected into the frozen DiT through a ControlNet branch. On the right, two object-centric attention (OCA) mechanisms impose complementary constraints in canonical and world spaces to maintain object-level consistency.}
\label{fig:method-overview}
\end{figure*}

\subsection{Object-centric 3D memory}
\label{sec:object-centric-memory}

Scene-level memory methods provide useful view-aligned guidance~\citep{yu2024viewcrafter,cao2025uni3c,ren2025gen3c,huang2025voyager,zhao2025spatia,zhang2026worldstereo}, but they do not maintain an object-centric 3D state that can be related in a shared world frame, making it difficult to preserve object-level consistency. Totem therefore constructs a persistent object-centric 3D memory set from \(I_0\) before video generation.

We first identify selected objects and obtain their instance masks from the reference image. Each masked object is lifted into an independent canonical 3D representation \(G_i\), which serves as a renderable carrier of appearance and geometry. Subsequently, we also estimate reference-view depth and define a shared world coordinate system \(W\). Each canonical object is then aligned to this world frame by an object-to-world Sim(3) transform \(S_i^{O_i \rightarrow W}\), bringing object memories, scene geometry, and target cameras into a common spatial reference. Finally, we sample canonical surface anchors \(\mathcal{A}_i\) from each object surface with farthest point sampling (FPS)~\citep{qi2017pointnetplusplus} and assign a stable object index.

The resulting memory set is
\begin{equation}
\mathcal{M}=\{\mathcal{M}_i\}_{i=1}^{N}, \qquad \mathcal{M}_i=(G_i,\mathcal{A}_i,S_i^{O_i \rightarrow W}).
\label{eq:objectmem3d}
\end{equation}
Here, \(G_i\) is the reconstructed 3D Gaussian of object \(i\)~\citep{chen2025sam3d}, and \(\mathcal{A}_i\) is a set of surface anchors sampled from the reconstructed Gaussian representation. \(S_i^{O_i \rightarrow W}\) maps a canonical object point \(x\in O_i\) into the shared world frame \(W\):
\[
S_i^{O_i \rightarrow W}(x)=s_iR_ix+t_i,
\]
where \(s_i>0\), \(R_i\in SO(3)\), and \(t_i\in\mathbb{R}^3\) denote the object scale, rotation, and translation.

\subsection{Dual-stream view guidance}
\label{sec:view-guidance}

With the memory set \(\mathcal{M}\), we render object-centric view conditions from the world-aligned 3D object representations stored in the memories. To provide world-level spatial context, scene depth is further used as a compact reference, together forming dual-stream view guidance. Specifically, we first place all aligned object memories into the world frame according to their object-to-world transforms \(S_i^{O_i \rightarrow W}\), compose them into a unified object representation, and render it from each target view. To handle occlusions by the background, we additionally generate a background-only depth map. Object pixels whose depths are larger than the corresponding background depths are filtered out from the object-memory RGB rendering and its mask, leaving only visible object regions. This produces two aligned condition streams,
\begin{equation}
\begin{aligned}
\mathcal{X}_{\mathrm{mem}}
&=\{(I_t^{\mathrm{mem}},M_t^{\mathrm{mem}})\}_{t=0}^{T-1},\\
\mathcal{X}_{\mathrm{depth}}
&=\{(D_t^{\mathrm{scene}},M_t^{\mathrm{scene}})\}_{t=0}^{T-1}.
\end{aligned}
\end{equation}
Here, \(\mathcal{X}_{\mathrm{mem}}\) and \(\mathcal{X}_{\mathrm{depth}}\) respectively provide per-view object RGB renderings and scene depth maps, with their corresponding masks. In parallel, \(O_t\) records the source object of each visible memory pixel, while \(X_t^{\mathrm{can}}\) stores its canonical surface coordinate, enabling surface binding.

We next inject these view conditions into a video prior, using a pretrained image-to-video diffusion transformer~\citep{wan2025wan,wanai2025wan22i2vdiffusers} as the generation backbone while keeping it frozen. Camera parameters are represented as Plücker rays and processed by a dedicated branch, with the resulting features incorporated through residual modulation to complement the rendered 3D conditions with trajectory information.  

For the dual-stream condition, we first encode the RGB sequence with the frozen video VAE
\((I^{\mathrm{mem}}\mapsto z^{\mathrm{mem}})\), while the depth sequence is log-normalized over valid pixels
as detailed in Appendix~\ref{app:model-architecture}, zero-filled elsewhere, replicated to three channels, and encoded by the same VAE
\((D^{\mathrm{scene}}\mapsto z^{\mathrm{depth}})\). The object and scene masks are independently max-pooled to the
latent resolution
\((M^{\mathrm{mem}}\mapsto m^{\mathrm{mem}},\, M^{\mathrm{scene}}\mapsto m^{\mathrm{depth}})\).

Because the RGB and depth have different statistical properties and semantic roles, each encoded condition is concatenated with its pooled mask and passed through a condition-specific learnable projection stem, \(P_{\mathrm{mem}}\) or \(P_{\mathrm{depth}}\), respectively, to map it to a common intermediate feature dimension. The resulting streams are then concatenated along the channel dimension and patchified to form control tokens:
\begin{equation}
\begin{aligned}
H_c
&=f_{\mathrm{emb}}^{c}
\Bigl(
\operatorname{concat}
\Bigl[
P_{\mathrm{mem}}
\Bigl(
\operatorname{concat}
[z^{\mathrm{mem}},m^{\mathrm{mem}}]
\Bigr),\\
&\qquad
P_{\mathrm{depth}}
\Bigl(
\operatorname{concat}
[z^{\mathrm{depth}},m^{\mathrm{depth}}]
\Bigr)
\Bigr]
\Bigr),
\end{aligned}
\label{eq:condition-fusion}
\end{equation}

In parallel, the noisy video latent is embedded through the original video pathway. The control and video tokens are combined by element-wise addition and processed by the ControlNet-style branch~\citep{zhang2023controlnet} associated with the active denoising expert, whose layer-wise residuals are injected into the corresponding blocks of the frozen video backbone through zero-initialized projections.

\subsection{Object-centric attention}
\label{sec:object-centric-attention}
\label{sec:canonical-surface-binding}
\label{sec:world-relation-preservation}

Although view guidance specifies what should appear at each target viewpoint, it does not explicitly model object states in canonical and world spaces. In more challenging scenarios, the same object may be regenerated independently across frames, leading to surface texture drift and changes in relative object positions. To address this limitation, we impose 3D constraints derived from object memory on attention in the ControlNet branch within these two complementary spaces. In
canonical space, visible tokens are bound to canonical surface anchors and refined through attention,
allowing surface information to persist. In world space, relative displacement, rotation,
and scale extracted from object-to-world transforms are converted into relation-aware attention biases,
helping preserve inter-object layout under camera motion.

\noindent\textbf{Canonical-space surface binding.}
For token \(n\) in frame \(t\), let \(\Pi_{t,n}\) denote its covered pixel region. We infer its object identity from the pixel support ratio of each object region within \(\Pi_{t,n}\), and discard tokens with weak support or mixed assignment. For a kept token with \(O_{t,n}=i\), we compute its canonical surface point \(x_{t,n}^{\mathrm{can}}\) by averaging \(X_t^{\mathrm{can}}\) over valid memory pixels assigned to object \(i\) within \(\Pi_{t,n}\). Its world coordinate is obtained using the object-to-world transform in the memory, \(p_{t,n}^{W}=S_i^{O_i\rightarrow W}(x_{t,n}^{\mathrm{can}})\).

For each visible token \(n\) assigned to object \(i\), we compute a correspondence weight to each anchor \(a_{i,m}\in\mathcal{A}_i\) based on the canonical-space distance between \(x_{t,n}^{\mathrm{can}}\) and \(a_{i,m}\):
\begin{equation}
w_{t,n,m}^{\mathrm{geo}}
=
\operatorname{softmax}_{m}
\left(
-\frac{\|x_{t,n}^{\mathrm{can}}-a_{i,m}\|_2^2}{\tau_x}
\right).
\label{eq:anchor-association}
\end{equation}
Here, \(\tau_x\) controls the scale of the canonical-coordinate distance, and the weights are normalized over anchors of object \(i\).

At layer \(l\), each anchor \(a_{i,m}\) aggregates the visible tokens assigned to object \(i\) via weighted averaging to form an anchor-level feature:
\begin{equation}
s_{t,i,m}^{(l)}
=
\frac{
\sum_n
\mathbf{1}[O_{t,n}=i]\,
w_{t,n,m}^{\mathrm{geo}}\,
z_{t,n}^{(l)}
}{
\sum_n
\mathbf{1}[O_{t,n}=i]\,
w_{t,n,m}^{\mathrm{geo}}
}.
\label{eq:surface-aggregation}
\end{equation}

\(s_{t,i,m}^{(l)}\) represents the anchor feature of \(a_{i,m}\) in frame \(t\). For each object-anchor pair \((i,m)\), canonical-space surface binding first binds the features corresponding to the same anchor across visible frames, then applies multi-head attention among these features, and finally writes the updated anchor state back to the associated tokens through a zero-initialized projection. In this way, anchors provide stable surface addresses, while their features carry surface states, enabling cross-frame information sharing and reducing surface drift.

\noindent\textbf{World-space relation preservation.}
For query object \(i\) and source object \(j\), we define
\begin{equation}
\begin{aligned}
S_{i\leftarrow j}
&=\left(S_i^{O_i\rightarrow W}\right)^{-1}\circ S_j^{O_j\rightarrow W},\\
g_{i\leftarrow j}
&=\bigl[\frac{t_{i\leftarrow j}}{s_i},R_{i\leftarrow j}^{6\mathrm{D}},\log s_{i\leftarrow j}\bigr].
\end{aligned}
\label{eq:relative-sim3}
\end{equation}
Here, \(t_{i\leftarrow j}\), \(R_{i\leftarrow j}^{6\mathrm{D}}\), and \(s_{i\leftarrow j}\) denote the relative displacement, rotation, and scale between object \(j\) and object \(i\). We normalize translation by the query object's scale \(s_i\), represent rotation with the continuous 6D parameterization, and encode relative scale in log space. The transform \(S_{i\leftarrow j}\) represents source object \(j\) in the coordinate frame of query object \(i\), making the relation directed.

For visible tokens \(r\) and \(s\) in the same frame with object IDs \(i\) and \(j\), \(i\neq j\), we compute token-level world points \(p_{t,r}^{W}\) and \(p_{t,s}^{W}\) from their canonical surface coordinates and use them to construct a geometric relation feature for biasing cross-object attention:
\begin{equation}
\begin{gathered}
e_{t,r,s}
=\operatorname{concat}\bigl[g_{i\leftarrow j},p_{t,s}^{W}-p_{t,r}^{W}\bigr],\\
\ell_{t,r,s}^{(l,h)}
=\frac{\left(q_{t,r}^{(l,h)}\right)^{\top}k_{t,s}^{(l,h)}}{\sqrt{d_h}}
 +\lambda_{\mathrm{rel}}^{(l,h)}\operatorname{MLP}_{\mathrm{rel}}^{(l,h)}\left(e_{t,r,s}\right).
\end{gathered}
\label{eq:relative-transform-attention}
\end{equation}
Here, \(e_{t,r,s}\) combines the object-level relative Sim(3) with the local 3D displacement between the two visible surface points. The learnable \(\lambda_{\mathrm{rel}}^{(l,h)}\) scales each layer-head bias, and the output is written back to control features through a zero-initialized projection. This relation modeling provides attention with stable object-layout information and supports relative transform consistency across objects.

%% file: sec/experiments.tex
\section{Experiments}

\subsection{Experimental setup}
\label{sec:data-engine}

\noindent\textbf{Implementation details.}
For 3D memory construction, we use Qwen3.5-122B-A10B~\citep{qwen2026qwen35} to identify the primary rigid objects and extract their boxes, SAM3~\citep{meta2025sam3} to obtain masks, and VGGT-Omega~\citep{wang2026vggtomega} to reconstruct a reference-view scene pointmap from the first frame. Each masked object is reconstructed as a canonical Gaussian with SAM3D~\citep{chen2025sam3d}, aligned to the scene with RANSAC~\citep{fischler1981ransac} and scale-constrained Umeyama initialization~\citep{umeyama1991least}, and then sampled into surface anchors with farthest point sampling (FPS)~\citep{qi2017pointnetplusplus}. For training, we initialize the frozen video backbone from the Wan2.2-I2V-A14B checkpoint~\citep{wan2025wan,wanai2025wan22i2vdiffusers,huggingface2026diffusers} and adopt a two-stage optimization schedule. We first train the camera and dual-stream control pathways, and then freeze the camera branch and enable the object-centric attention modules. Complete training and inference settings are provided in Appendix~\ref{app:training-inference}.

\noindent\textbf{Training data.}
We train on two open-source real-world datasets, RealEstate10K~\citep{zhou2018stereo} and DL3DV~\citep{ling2024dl3dv}. RealEstate10K mainly contains real-estate videos with indoor residential scenes and some outdoor environments, while DL3DV provides diverse real-scene videos with richer camera motion. After filtering unsuitable samples and constructing fixed-length windows, we retain about 40K clips from each dataset, resulting in roughly 80K training clips. All clips are resampled to 15 fps, split into 81-frame windows, and center-cropped to 720p.

\noindent\textbf{Evaluation.}
We evaluate Totem from two aspects: (1) world consistency and (2) object consistency. For (1), we adopt WorldScore~\citep{duan2025worldscore}, which consists of 2,000 static test examples spanning diverse worlds, including indoor and outdoor, photorealistic and stylized scenes. Each example provides an input image and a camera trajectory. The metrics evaluate generation controllability and quality, and the overall score summarizes the final performance. For (2), the public subset of ConsIDVid-Bench~\citep{wu2026considgen} is selected to evaluate object consistency. It is an object-centric video benchmark with 370 videos drawn from public datasets and synthetic sequences that evaluates both perceptual and geometric consistency at the object level.

\begin{table*}[t]
\centering
\caption{Quantitative comparison on WorldScore Benchmark. \textbf{Bold} indicates the best result; underlining indicates the second-best result.}
\label{tab:worldscore_static}
\fontsize{10pt}{12pt}\selectfont
\begin{equalspacedtabular}{8}{lG{0.8801pt}{0.0499pt}cG{0.2000pt}{0.2800pt}cS{0.3001pt}cS{0.3601pt}cG{-0.0099pt}{0.2800pt}cS{0.4101pt}cS{0.6001pt}cS{0.6700pt}c}
\toprule
Method & \tablehead{WorldScore\\Static} & \tablehead{Camera\\Control} & \tablehead{Object\\Control} & \tablehead{Content\\Alignment} & \tablehead{3D\\Consistency} & \tablehead{Photometric\\Consistency} & \tablehead{Style\\Consistency} & \tablehead{Subjective\\Quality} \\
\midrule
Voyager~\citep{huang2025voyager} & 77.62 & 85.95 & 66.92 & 68.92 & 81.56 & 85.99 & 84.89 & \textbf{71.09} \\
Spatia~\citep{zhao2025spatia} & 72.63 & 75.66 & 52.32 & 69.95 & 86.40 & 89.10 & 80.09 & 54.86 \\
\midrule
Mirage~\citep{wang2026mirage} & 73.60 & 55.36 & 74.17 & 42.09 & 92.21 & 93.95 & \textbf{96.91} & 60.50 \\
EonWorld~\citep{eonworld2026} & 81.08 & 79.51 & 61.00 & 82.12 & \underline{92.76} & 89.18 & \underline{96.70} & 66.28 \\
WorldScape-0.2~\citep{manifold2026worldscape} & 85.13 & 94.32 & 87.65 & 78.13 & 86.51 & 90.02 & 90.52 & \underline{68.75} \\
UniWorld-View~\citep{zhou2026uniworldviewlargebaselineviewsynthesis} & \underline{85.53} & \underline{97.72} & \textbf{88.98} & \textbf{86.61} & 91.63 & \underline{94.11} & 76.55 & 63.12 \\
\midrule
\textbf{Totem} & \textbf{86.43} & \textbf{98.31} & \underline{88.24} & \underline{84.52} & \textbf{93.12} & \textbf{95.25} & 79.43 & 66.15 \\
\bottomrule
\end{equalspacedtabular}
\end{table*}

\begin{table*}[t]
\centering
\caption{Quantitative comparison on ConsIDVid-Bench. \textbf{Bold} indicates the best result; underlining indicates the second-best result.}
\label{tab:considvid}
\small
\begin{equalspacedtabular}{10}{lG{0.7920pt}{0.3779pt}cS{0.1441pt}cS{0.3330pt}cS{0.3781pt}cS{0.4590pt}cS{0.4770pt}cG{0.2701pt}{0.3779pt}cS{0.6030pt}cG{0.2251pt}{0.1439pt}cS{0.0900pt}c}
\toprule
Method & \tablehead{I2V\\Subject} & \tablehead{I2V\\Background} & \tablehead{Subject\\Consistency} & \tablehead{Background\\Consistency} & \tablehead{Motion\\Smoothness} & \tablehead{Temporal\\Flickering} & \tablehead{Video\\Similarity} & \tablehead{Object\\Similarity} & \tablehead{Chamfer\\Distance} & MEt3R \\
\midrule
ViewCrafter~\citep{yu2024viewcrafter} & 95.73 & 95.58 & 92.81 & 93.74 & 97.64 & 96.28 & 81.47 & 67.4 & 0.1423 & 0.1308 \\
GEN3C~\citep{ren2025gen3c} & 96.38 & 97.42 & 93.86 & 95.33 & 98.57 & 97.52 & 83.92 & 70.3 & 0.1154 & 0.1057 \\
Voyager~\citep{huang2025voyager} & 96.72 & 97.18 & 93.28 & 95.06 & 98.17 & 96.85 & 83.24 & 69.5 & 0.1128 & 0.1103 \\
Spatia~\citep{zhao2025spatia} & \underline{97.34} & 97.63 & \underline{94.15} & 95.27 & \textbf{99.16} & \textbf{98.44} & \underline{84.56} & 71.1 & 0.1246 & 0.1176 \\
WorldStereo~\citep{zhang2026worldstereo} & 96.93 & \textbf{98.06} & 93.98 & \textbf{95.82} & 98.48 & 97.36 & 83.76 & \underline{71.4} & \underline{0.1053} & \underline{0.0964} \\
\midrule
\textbf{Totem} & \textbf{97.76} & \underline{97.86} & \textbf{95.04} & \underline{95.62} & \underline{98.82} & \underline{97.75} & \textbf{85.08} & \textbf{73.2} & \textbf{0.0976} & \textbf{0.0818} \\
\bottomrule
\end{equalspacedtabular}
\end{table*}

Additionally, to evaluate Totem's generalization in complex settings, we introduce an out-of-distribution (OOD) benchmark including 400 videos, constructed from RealEstate10K and DL3DV. The OOD benchmark is stratified into four object-centric challenge types, each with 100 clips: target-object disappearance due to occlusion or out-of-view motion, confusion with nearby or overlapping similar objects, revisited views after large viewpoint changes, and scale or resolution degradation under long-range camera motion.

To emphasize object-level validation, we first evaluate the selected objects using three metrics from ConsIDVid-Bench (Object Similarity, Chamfer Distance, and MEt3R). The masks for the selected objects are obtained by reapplying the mask-extraction stage of our memory construction pipeline to the generated videos. For Object Similarity, we follow the benchmark's DINO-based feature comparison, while matching each generated object to the reference observation with the closest camera viewpoint. For Chamfer Distance and MEt3R, complete frames are used to estimate multi-view geometry and feature correspondences, after which the same object masks are applied to retain object-specific points and features before aggregation. In addition, we report PSNR, SSIM, and LPIPS to measure overall visual fidelity.

\subsection{Main results}

\noindent\textbf{World consistency.}
We compare Totem with six top methods on the WorldScore benchmark, including two representative 3D memory methods, Voyager and Spatia~\citep{huang2025voyager,zhao2025spatia}, and four strong world-generation baselines, Mirage~\citep{wang2026mirage}, EonWorld, WorldScape-0.2~\citep{manifold2026worldscape}, and UniWorld-View~\citep{zhou2026uniworldviewlargebaselineviewsynthesis}. As reported in Table~\ref{tab:worldscore_static}, Totem achieves the highest WorldScore-Static score on this benchmark. In particular, it leads on camera control, 3D consistency, and photometric consistency. These results indicate that Totem provides effective object-level spatial support for world-consistent generation.

\noindent\textbf{Object consistency.}
We further compare Totem with five 3D scene-aware video generation baselines for object-centric consistency. The comparison set spans three representative paradigms: 3D-cache-guided view synthesis, including ViewCrafter and GEN3C~\citep{yu2024viewcrafter,ren2025gen3c}; persistent scene-level memory methods, including Voyager and Spatia~\citep{huang2025voyager,zhao2025spatia}; and geometry-memory hybrid generation, represented by WorldStereo~\citep{zhang2026worldstereo}, which combines camera-guided video synthesis with 3D geometric memory. As reported in Table~\ref{tab:considvid}, Totem achieves the best results across the object-centric metrics. Higher Object Similarity reflects better instance-specific appearance, while lower Chamfer Distance and MEt3R correspond to more stable shape and cross-view geometry.

\begin{figure*}[t]
\centering
\includegraphics[width=\linewidth]{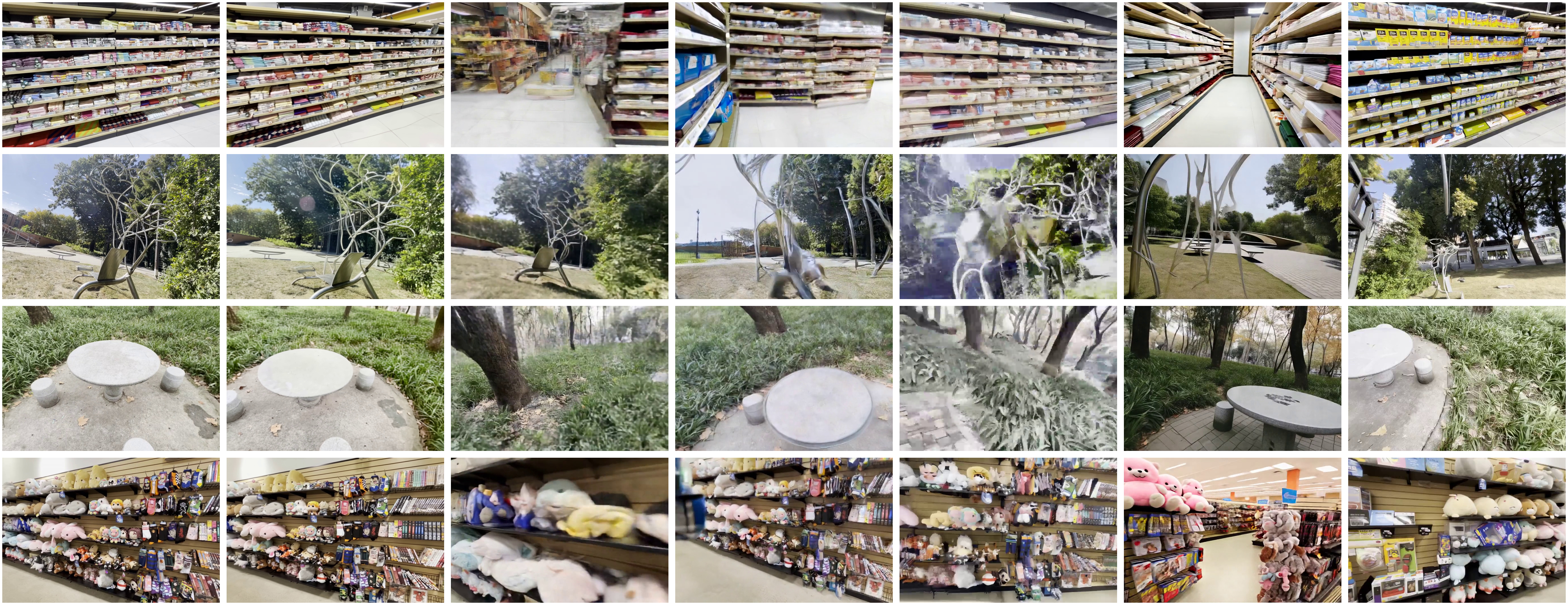}
\par\vspace{\floatcontentsep}
\paperfigurelabel{%
\noindent\hbox to \linewidth{%
\hspace*{0.001445\linewidth}%
\hbox to 0.138728\linewidth{\hfil GT\hfil}%
\hspace{0.004335\linewidth}%
\hbox to 0.138728\linewidth{\hfil Ours\hfil}%
\hspace{0.004335\linewidth}%
\hbox to 0.138728\linewidth{\hfil GEN3C\hfil}%
\hspace{0.004335\linewidth}%
\hbox to 0.138728\linewidth{\hfil Spatia\hfil}%
\hspace{0.004335\linewidth}%
\hbox to 0.138728\linewidth{\hfil ViewCrafter\hfil}%
\hspace{0.004335\linewidth}%
\hbox to 0.138728\linewidth{\hfil Voyager\hfil}%
\hspace{0.004335\linewidth}%
\hbox to 0.138728\linewidth{\hfil WorldStereo\hfil}%
\hspace*{0.001445\linewidth}%
\hss}}
\vspace{-9.6pt}
\caption{Qualitative results on video generation. Compared to the baselines, our model can generate videos with higher consistency while preserving the content in the input view.}
\label{fig:video-consistency}
\end{figure*}

\begin{table*}[t]
\centering
\caption{Ablation study of Dual-Stream View Guidance and Object-Centric Attention
(OCA) on the OOD benchmark.}
\label{tab:ablation}
\normalsize
\newcommand{\ablationtwolinemetrichead}[3]{\tablehead{#1\\#2}\,#3}
\begin{equalspacedtabular}[0.1702pt]{6}{lG{0.3359pt}{0.1663pt}cS{0.5987pt}cS{0.2986pt}cG{0.1760pt}{0.4264pt}cS{0.2751pt}cS{0.3031pt}c}
\toprule
Variant & \tablehead{PSNR\,\(\uparrow\)} &
\tablehead{SSIM\,\(\uparrow\)} &
\tablehead{LPIPS\,\(\downarrow\)} &
\ablationtwolinemetrichead{\kern-.104pt Object\kern.104pt}{Similarity}{\raisebox{-.192pt}[0pt][0pt]{\(\uparrow\)}} &
\raisebox{-1.01pt}{\ablationtwolinemetrichead{\kern-.1875pt Chamfer\kern.1875pt}{Distance}{\raisebox{.805pt}[0pt][0pt]{\(\downarrow\)}}} &
\tablehead{MEt3R\,\(\downarrow\)} \\
\midrule
Wan2.2~\citep{wan2025wan,wanai2025wan22i2vdiffusers} with Camera Branch & 17.31 & 0.528 & 0.466 & 48.5 & 0.289 & 0.361 \\
+ Projected Point Cloud & 21.04 & 0.653 & 0.347 & 53.7 & 0.234 & 0.284 \\
+ \mbox{Dual-Stream View Guidance} & 23.15 & 0.759 & 0.241 & 58.6 & 0.173 & 0.208 \\
\textbf{+ \mbox{Dual-Stream View Guidance + OCA}} & \textbf{25.67} & \textbf{0.838} & \textbf{0.139} & \textbf{65.3} & \textbf{0.124} & \textbf{0.138} \\
\bottomrule
\end{equalspacedtabular}
\end{table*}

\noindent\textbf{Visualization.}
Figure~\ref{fig:video-consistency} provides qualitative comparisons across diverse indoor and outdoor scenes. The results further support the quantitative findings above, showing strong consistency at both the world and object levels. In particular, in the second example, as the camera moves around the statue, only Totem preserves the statue's appearance and fine-grained details, whereas the competing methods tend to introduce visible artifacts, such as blurred textures and distorted structures.

\subsection{Ablation studies}

To verify the effectiveness of our proposed designs, we conduct ablation studies on dual-stream view guidance and object-centric attention. All variants are evaluated on the OOD benchmark under the same protocol, isolating the contribution of each component. Table~\ref{tab:ablation} compares four variants: (a) the original Wan2.2 I2V backbone augmented with the camera branch; (b) our model with object-memory and scene-depth dual-stream view guidance; (c) variant (a) conditioned on view-aligned projections of a scene point cloud reconstructed from the first frame, using the same injection scheme and training settings as (b); and (d) our full model with object-centric attention.

\noindent\textbf{Dual-stream view guidance.}
We first evaluate the effect of view guidance by comparing the first three variants. As shown in Table~\ref{tab:ablation}, introducing projected point-cloud guidance enhances generation quality over the camera-branch baseline, indicating the benefit of explicit 3D conditions. Replacing point-cloud projections with our dual-stream view guidance further improves all metrics, showing that our dual-stream view guidance leads to higher visual fidelity and stronger spatial coherence in the generated videos.

\noindent\textbf{Object-centric attention.}
We then examine the final step of adding object-centric attention to the dual-stream model. The gains across the object-centric metrics demonstrate that, beyond the initial 3D priors, object-centric attention brings additional gains in cross-view object appearance stability and geometric alignment under challenging camera trajectories.

\begin{figure*}[t]
    \centering
    \begin{minipage}[t]{0.48\textwidth}
        \vspace{0pt}
        \centering
        \parbox[t][2em][t]{\linewidth}{\centering\paperfigurelabel{(a) Camera-Controllable Video Generation}}\\[\floatcontentsep]
        \includegraphics[trim=0 0 0 39pt,clip,width=\linewidth]{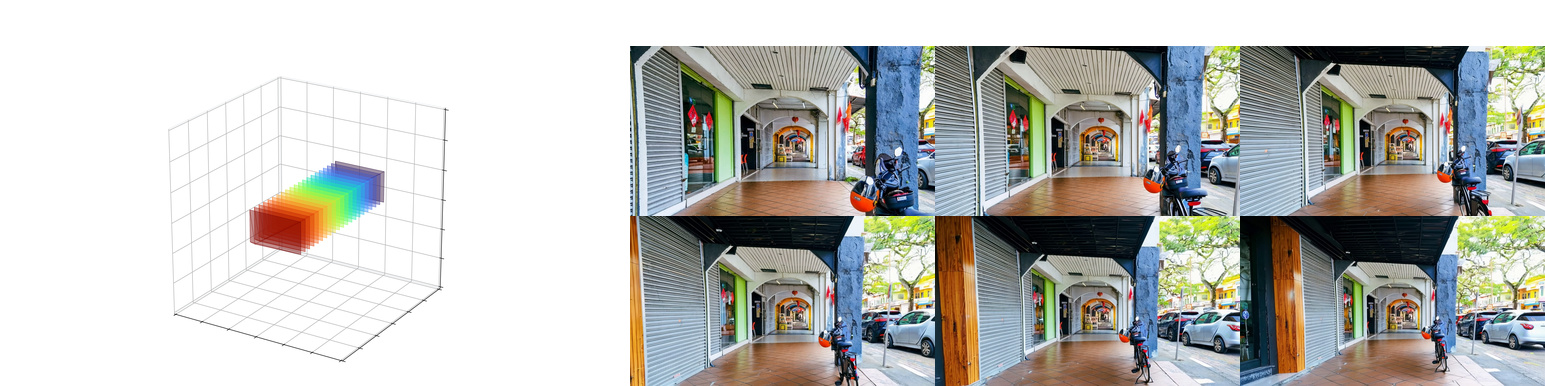}\\[\floatcontentsep]
        \includegraphics[trim=0 0 0 39pt,clip,width=\linewidth]{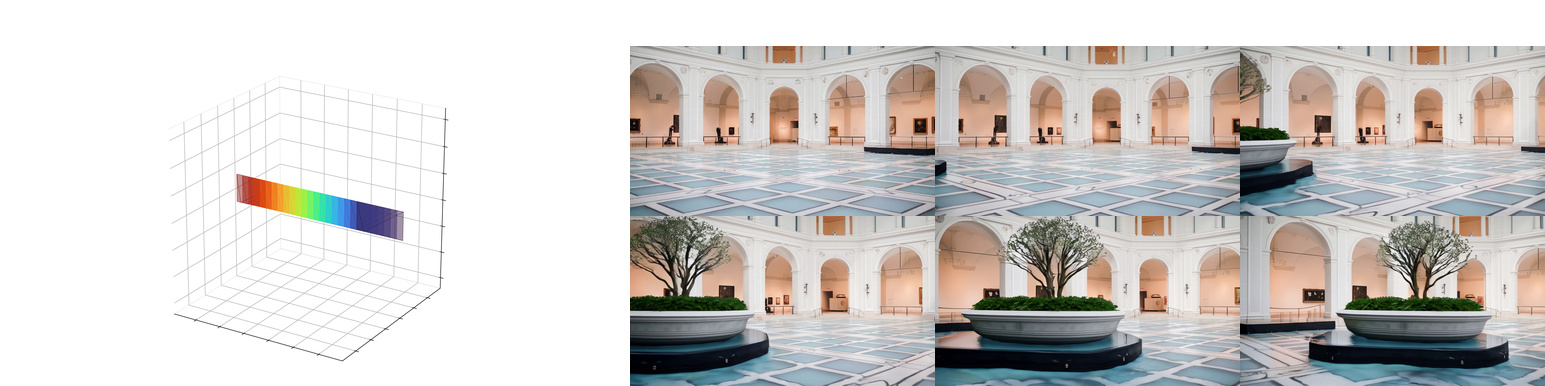}
    \end{minipage}
    \hfill
    \begin{minipage}[t]{0.48\textwidth}
        \vspace{0pt}
        \centering
        \parbox[t][2em][t]{\linewidth}{\centering\paperfigurelabel{(b) 3D-Aware Video Editing}}\\[\floatcontentsep]
        \includegraphics[trim=0 0 0 41pt,clip,width=\linewidth]{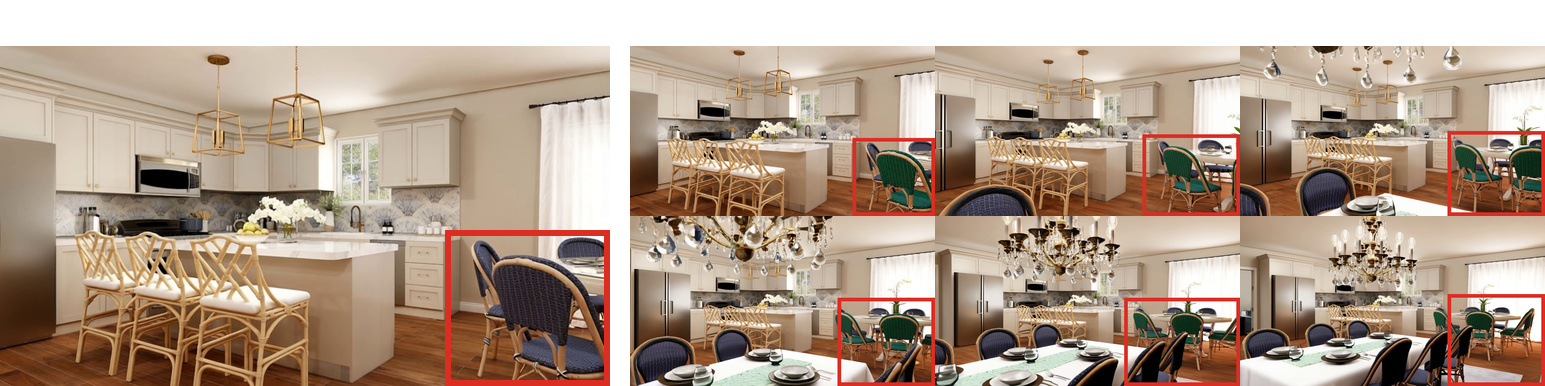}\\[\floatcontentsep]
        \includegraphics[trim=0 0 0 45pt,clip,width=\linewidth]{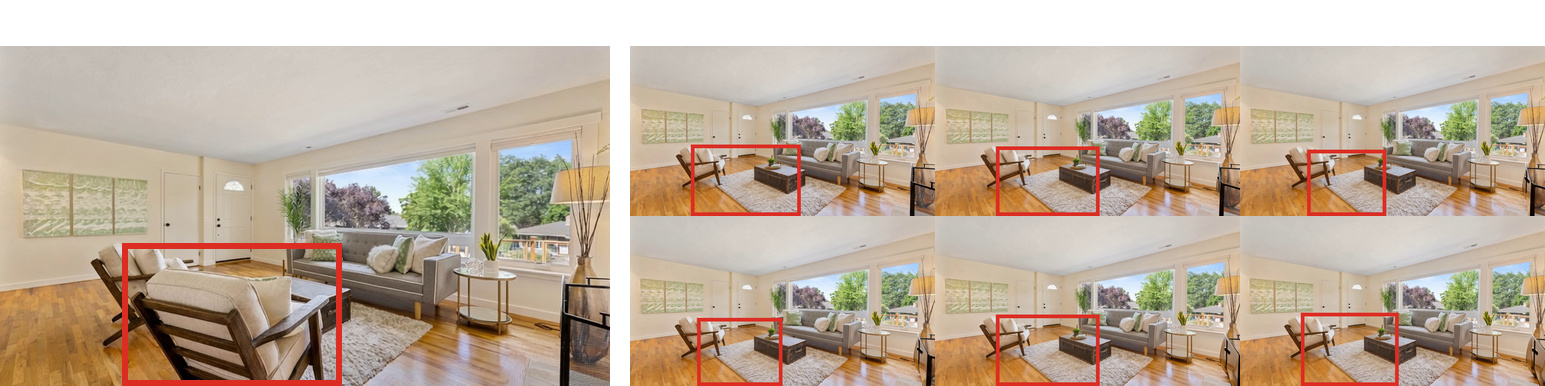}
    \end{minipage}
    \par\vspace{\dimexpr\abovecaptionskip+0.65pt\relax}
    \parbox[t][2em][t]{\textwidth}{\centering\paperfigurelabel{(c) Long-Range Video Generation}}\\[\floatcontentsep]
    \includegraphics[width=\textwidth]{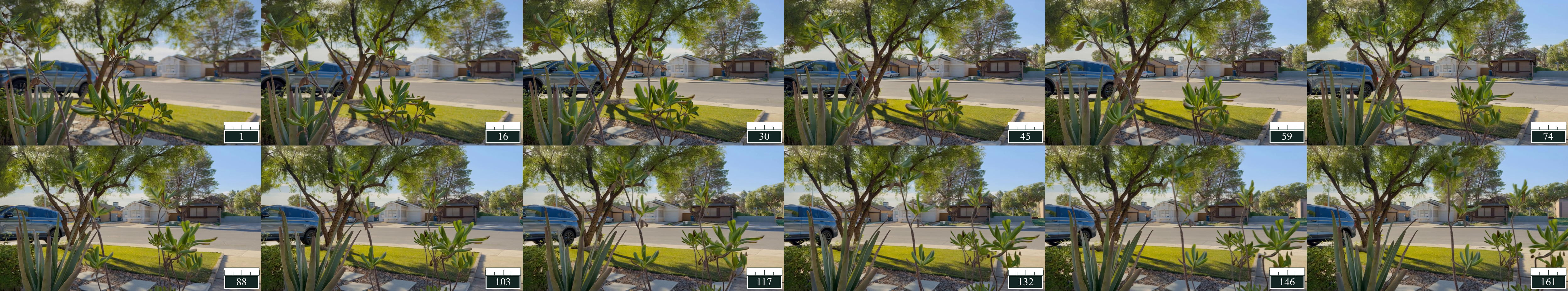}
    \caption{Applications: (a) Camera-controllable video generation. (b) 3D-aware video editing. (c) Long-range video generation.}
    \label{fig:applications}
\end{figure*}

\begin{table*}[t]
\centering
\caption{Quantitative comparison of camera-controllable video generation on the OOD
benchmark. Translation-related errors are computed on scale-normalized trajectories, RPE is
measured between consecutive frames, and ATE is calculated after global Sim(3)
alignment.}
\label{tab:camera_control}
\normalsize
\begin{equalspacedtabular}[0.1660pt]{5}{lG{0.8788pt}{0.1690pt}cS{0.3396pt}cG{0.1720pt}{0.1700pt}cS{0.3396pt}cG{0.1715pt}{0.1500pt}c}
\toprule
Method & \tablehead{RotErr\,\(\downarrow\)} & \tablehead{TransErr\,\(\downarrow\)} & \tablehead{RPE-T\,\(\downarrow\)} & \tablehead{RPE-R\,\(\downarrow\)} & \tablehead{ATE\,\(\downarrow\)} \\
\midrule
AC3D~\citep{bahmani2024ac3d} & 16.87 & 0.358 & 4.12 & 1.58 & 0.842 \\
CamI2V~\citep{zheng2024cami2v} & 14.63 & 0.312 & 3.34 & 1.42 & 0.689 \\
CameraNoise~\citep{zhao2026cameranoise} & 8.92 & 0.167 & 2.23 & 0.683 & 0.388 \\
RealCam-I2V~\citep{li2025realcam} & 8.14 & 0.134 & 2.41 & 0.911 & 0.329 \\
TriMotion-Pose~\citep{shin2026trimotion} & 5.38 & 0.146 & 1.57 & 0.462 & 0.253 \\
\midrule
\textbf{Totem} & \textbf{3.82} & \textbf{0.077} & \textbf{1.06} & \textbf{0.418} & \textbf{0.148} \\
\bottomrule
\end{equalspacedtabular}
\end{table*}

%% file: sec/applications.tex
\section{Applications}

By leveraging its object-centric 3D memories, Totem supports various video-related applications.

\noindent\textbf{Camera-controllable video generation.}
As explained in Sec.~\ref{sec:view-guidance}, our dual-stream conditions provide
view-aligned guidance for both scene geometry and object appearance
at each target viewpoint, thereby enabling camera-controllable video
generation. In Table~\ref{tab:camera_control}, we compare
Totem with five mainstream image-to-video methods with camera conditions~\citep{bahmani2024ac3d,zheng2024cami2v,li2025realcam,
zhao2026cameranoise,shin2026trimotion} on the OOD benchmark. We
evaluate camera controllability with absolute pose errors (RotErr and
TransErr), local motion errors (RPE-T and RPE-R), and global trajectory
error (ATE). The results demonstrate that Totem achieves stronger camera
controllability and spatial stability in the generated videos. Figure~\ref{fig:applications}(a) further presents two examples with the Move Right and Push In camera motions. In both cases, the high-quality generated frames follow the predefined camera trajectories.

\noindent\textbf{3D-aware video editing.}
Having demonstrated the effectiveness of object-centric 3D memories
for consistent video synthesis, we further extend Totem to object-level
video editing. Given a reference image, users edit the selected object
in the first frame using an image-editing model. The corresponding 3D memory is then updated and used to generate the edited video. In Figure~\ref{fig:applications}(b), we provide two examples of object change and object removal, where Totem successfully propagates these edits across viewpoints, producing visually coherent videos while preserving the surrounding scene and the spatial relationships among unedited objects.

\noindent\textbf{Long-range video generation.}
Totem also supports long-range generation through autoregressive rollout.
We obtain object representations from each clip's terminal frame and add
them to the object-centric 3D memory to guide the next clip. This extends
generation beyond a single clip for sustained world exploration.
Figure~\ref{fig:applications}(c) shows camera motion continuing across
clips and revealing new outdoor views.

%% file: sec/conclusion.tex
\section{Conclusion}

In this paper, we present Totem, an object-centric 3D memory framework
for world-consistent video generation. Totem generates coherent
videos with stable spatial structure and object appearance across different
viewpoints. This capability further supports controllable, editable, and long-horizon video generation. Extensive experiments demonstrate improved visual
fidelity, world consistency, and object consistency under
challenging camera trajectories.

%% file: sec/supplementary_material.tex

\section{Object-centric 3D memory construction}
\label{app:memory-construction}

\subsection{Object-to-world alignment}
\label{app:object-world-alignment}

\noindent\textbf{Coordinate convention.}
We use the reference-camera coordinates of the VGGT-Omega~\citep{wang2026vggtomega}
pointmap \(P_0^W\) as the world frame \(W\). Each SAM3D~\citep{chen2025sam3d}
Gaussian \(G_i\) lies in an object-specific
canonical frame \(O_i\) with unknown metric scale. We estimate its object-to-world
Sim(3) transform by global initialization followed by scale-locked refinement
against reference-view observations.

\noindent\textbf{Pixel-aligned correspondences.}
The reference image, mask, pointmap, and SAM3D source-view render are aligned
using the same image transform. We erode the object mask to reduce boundary
contamination and render the object to obtain canonical coordinates, alpha, and
depth. We retain pixels where both the object render and VGGT point are valid
and the rendered object is frontmost, giving the tentative 3D--3D pairs
\begin{equation}
\mathcal{C}_i
=
\left\{
\left(X_{i,0}^{\mathrm{can}}(u),P_0^W(u)\right)
\;\middle|\;
u\in\Omega_i^{\mathrm{valid}}
\right\}.
\label{eq:appendix-correspondences}
\end{equation}

\noindent\textbf{Robust initialization and refinement.}
For a point set \(Y\), define
\(\Delta_{5:95}(Y)=\lVert\mathbf{q}_{0.95}(Y)-\mathbf{q}_{0.05}(Y)\rVert_2\),
where the quantiles are taken coordinate-wise. Let
\(\mathcal{P}_i^{\mathrm{mask}}\) denote the valid VGGT points inside the
eroded reference mask, and let \(\mathcal{P}_i=\{P_0^W(u)\}_{u\in\Omega_i^{\mathrm{valid}}}\)
and \(\mathcal{X}_i=\{X_{i,0}^{\mathrm{can}}(u)\}_{u\in\Omega_i^{\mathrm{valid}}}\)
denote the pixel-aligned point sets. We use a two-branch initialization strategy:
a pixel-aligned main branch is applied when reliable correspondences are available,
while a normalized FPFH-based branch~\citep{rusu2009fast} serves as a fallback
when the main branch is unavailable or unreliable. For the main branch, we use
RANSAC~\citep{fischler1981ransac} followed by closed-form Umeyama estimation~\citep{umeyama1991least}.
Its scale prior and the observed object size
used to normalize distance thresholds are
\begin{equation}
s_i^{(0)}=\frac{\Delta_{5:95}(\mathcal{P}_i)}{\Delta_{5:95}(\mathcal{X}_i)+\epsilon},
\qquad d_i=\Delta_{5:95}(\mathcal{P}_i^{\mathrm{mask}}).
\label{eq:appendix-scale-prior}
\end{equation}
For a RANSAC inlier set \(\mathcal{I}_i\), write
\(T_{s,R,t}(x)=sRx+t\) and define the residual
\(r_u(s,R,t)=P_0^W(u)-T_{s,R,t}\!\left(X_{i,0}^{\mathrm{can}}(u)\right)\).
The Umeyama estimate is
\begin{equation}
(\hat{s}_i,\hat{R}_i,\hat{t}_i)
=
\arg\min_{s,R,t}
\sum_{u\in\mathcal{I}_i}\|r_u(s,R,t)\|_2^2,
\label{eq:appendix-sim3-init}
\end{equation}
and we retain only hypotheses satisfying
\begin{equation}
\frac{s_i^{(0)}}{\gamma}\leq \hat{s}_i\leq\gamma s_i^{(0)}.
\label{eq:appendix-scale-acceptance}
\end{equation}
Here, \(\gamma>1\) controls the admissible scale range; hypotheses outside it
are rejected. Main-branch initialization requires sufficient pixel support and
an inlier ratio above its threshold (Table~\ref{tab:memory-parameters}).

If pixel support is insufficient, the main-branch inlier ratio fails its
threshold, or source-view visibility fails, we invoke the FPFH fallback. It
matches the valid masked VGGT point cloud to canonical surface points
\(\mathcal{X}_i^{\mathrm{surf}}\) sampled directly from the Gaussian geometry,
without pixel-aligned source-view correspondences. The fallback uses an
independent scale prior:
\begin{equation}
s_{i,\mathrm{FPFH}}^{(0)}
=\frac{\Delta_{5:95}(\mathcal{P}_i^{\mathrm{mask}})}
{\Delta_{5:95}(\mathcal{X}_i^{\mathrm{surf}})+\epsilon}.
\label{eq:appendix-fpfh-scale-prior}
\end{equation}
The fallback is rejected if either point set lacks sufficient support. Its
inlier ratio is computed over tentative FPFH matches; the inlier-ratio and
distance thresholds are listed in Table~\ref{tab:memory-parameters}. Its scale
must satisfy
\(s_{i,\mathrm{FPFH}}^{(0)}/\gamma \leq \hat{s}_i \leq
\gamma s_{i,\mathrm{FPFH}}^{(0)}\). If neither branch returns a valid
initialization, the object is discarded; otherwise we set
\((s_i,R_i,t_i)=(\hat{s}_i,\hat{R}_i,\hat{t}_i)\).

After global initialization, we reproject the complete object into the reference
view. A surface point is observation-supported only if it projects inside the
safe eroded object mask, has positive camera depth, is frontmost in the object
depth buffer, and satisfies
\(\lvert D_i^{\mathrm{obj}}(u)-D_0^W(u)\rvert\leq\tau_{\mathrm{obs}}d_i\),
with \(\tau_{\mathrm{obs}}=0.03\). This post-initialization filtering prevents
unobserved backsides and unsupported single-view completions from entering
refinement. We lock the accepted scale and optimize only a rotation and
translation residual using
\begin{equation}
\begin{aligned}
\mathcal{L}_{\mathrm{align}}
&=\lambda_{\mathrm{3d}}\mathcal{L}_{\mathrm{p2plane}}
 +\lambda_d\mathcal{L}_{\mathrm{depth}}\\
&\quad +\lambda_b\mathcal{L}_{\mathrm{boundary}}
 +\lambda_m\mathcal{L}_{\mathrm{mask}},
\qquad s_i\ \text{fixed}.
\end{aligned}
\label{eq:appendix-alignment-loss}
\end{equation}
The four terms measure point-to-plane, depth, projected-boundary, and mask
agreement on reference-view observations.

\noindent\textbf{Acceptance and failure handling.}
After refinement, both branches use the same reference-view validation checks;
their thresholds are listed in Table~\ref{tab:memory-parameters}. Only
alignments that pass all checks are retained in the memory set.

\subsection{Construction algorithm}
\label{app:construction-algorithm}

Algorithm~\ref{alg:memory-construction} summarizes the complete memory construction procedure, which is performed before video denoising. In addition to the object RGB and scene-depth streams consumed by the dual-stream ControlNet, it retains object identities, canonical coordinates, and object-to-world transforms as geometric metadata for object-centric attention.

\begin{algorithm*}[t]
\caption{Memory Construction Pipeline}
\label{alg:memory-construction}
\normalsize
\begin{algorithmic}[1]
\REQUIRE Reference image \(I_0\); target cameras \(\{C_t,K_t\}_{t=0}^{T-1}\)
\ENSURE Memory \(\mathcal{M}\); conditions \(\mathcal{X}_{\mathrm{mem}},\mathcal{X}_{\mathrm{depth}}\); metadata \(\{O_t,X_t^{\mathrm{can}}\}_{t=0}^{T-1}\)
\STATE Estimate the reference pointmap \(P_0^W\) with VGGT-Omega
\STATE Detect object boxes and obtain masks \(\{M_i^0\}_{i=1}^{N}\)
\STATE Initialize \(\mathcal{M}\leftarrow\varnothing\)
\STATE Initialize the accepted-mask set \(\mathcal{S}_{\mathrm{acc}}\leftarrow\varnothing\)
\FOR{each selected object \(i\)}
    \STATE Reconstruct its canonical Gaussian representation \(G_i\) with SAM3D
    \STATE Render source-view canonical coordinates, alpha, and depth
    \STATE Construct pixel-aligned correspondences \(\mathcal{C}_i\)
    \IF{at least 128 correspondences and source-view visibility passes}
        \STATE Estimate \(S_i^{O_i\rightarrow W}\) by scale-guarded RANSAC--Umeyama
    \ELSE
        \STATE Mark the main initialization as unavailable
    \ENDIF
    \IF{main initialization is unavailable or its inlier ratio is below 0.25}
        \STATE Re-estimate \(S_i^{O_i\rightarrow W}\) with normalized FPFH and Sim(3) RANSAC
        \STATE Accept fallback initialization only if its FPFH inlier ratio is at least 0.20 and its scale is valid
    \ENDIF
    \IF{neither initialization branch returns a valid Sim(3) hypothesis}
        \STATE Discard object \(i\) and continue
    \ENDIF
    \STATE Filter surfaces without reference-view observation support
    \STATE Refine rotation and translation while keeping the scale fixed
    \IF{the alignment fails the acceptance checks}
        \STATE Discard object \(i\) and continue
    \ELSE
        \STATE Sample canonical surface anchors \(\mathcal{A}_i\) using FPS
        \STATE Add \((G_i,\mathcal{A}_i,S_i^{O_i\rightarrow W})\) to \(\mathcal{M}\) and \(M_i^0\) to \(\mathcal{S}_{\mathrm{acc}}\)
    \ENDIF
\ENDFOR
\STATE Remove only masks in \(\mathcal{S}_{\mathrm{acc}}\) from \(P_0^W\) to obtain \(P_{\mathrm{bg}}^W\); rejected candidates remain available as scene geometry
\FOR{\(t=0,\ldots,T-1\)}
    \STATE Jointly render all aligned memories under \(C_t\) with a shared depth buffer
    \STATE Obtain \(I_t^{\mathrm{obj}},\alpha_t^{\mathrm{obj}},D_t^{\mathrm{obj}},O_t,X_t^{\mathrm{can}}\)
    \STATE Render \(P_{\mathrm{bg}}^W\) to obtain \(D_t^{\mathrm{scene}}\) and its valid mask
    \STATE Resolve object--scene visibility using a common camera-depth convention
    \STATE Store \((I_t^{\mathrm{mem}},M_t^{\mathrm{mem}})\) and \((D_t^{\mathrm{scene}},M_t^{\mathrm{scene}})\)
\ENDFOR
\RETURN \(\mathcal{M},\mathcal{X}_{\mathrm{mem}},\mathcal{X}_{\mathrm{depth}},\{O_t,X_t^{\mathrm{can}}\}_{t=0}^{T-1}\)
\end{algorithmic}
\end{algorithm*}

\begin{table*}[!t]
\centering
\caption{Memory construction parameters.}
\label{tab:memory-parameters}
\normalsize
\begin{tabularx}{\textwidth}{@{}>{\raggedright\arraybackslash}p{0.34\textwidth}>{\raggedright\arraybackslash}X@{}}
\toprule
Parameter & Setting \\
\midrule
\rowcolor{TableGroup}[0pt][0pt]
\multicolumn{2}{@{}l@{}}{\textit{Preprocessing}} \\
Mask erosion & 3-pixel radius at 512-pixel resolution \\
Point-cloud preprocessing & Voxel size \(0.01d_i\); 30-neighbor normal estimation within radius \(0.03d_i\) \\
\midrule
\rowcolor{TableGroup}[0pt][0pt]
\multicolumn{2}{@{}l@{}}{\textit{Robust initialization}} \\
Pixel correspondence test & At least 128 pairs; alpha above 0.05 and positive camera depth \\
RANSAC--Umeyama & 4,096 iterations; 3-point hypotheses; inlier distance \(0.02d_i\); inlier ratio at least 0.25 \\
Scale prior & Ratio of the 5th--95th percentile 3D extents; \(\gamma=2\) in Eq.~\eqref{eq:appendix-scale-acceptance} \\
FPFH fallback & Voxel size \(0.02d_i\); normal radius \(3v\); feature radius \(5v\); 8,192 RANSAC iterations; inlier distance \(0.05d_i\); inlier ratio \(\geq0.20\) \\
\midrule
\rowcolor{TableGroup}[0pt][0pt]
\multicolumn{2}{@{}l@{}}{\textit{Refinement and validation}} \\
Local refinement & 100 Adam steps on an \(\mathrm{SE}(3)\) residual; learning rate \(10^{-3}\); \(\beta_1=0.9\), \(\beta_2=0.999\), \(\epsilon=10^{-8}\); fixed scale \\
Alignment loss weights & \(\lambda_{\mathrm{3d}}=1\), \(\lambda_d=1\), \(\lambda_b=0.1\), \(\lambda_m=0.5\) \\
Alignment acceptance & Mask IoU \(\geq0.50\); boundary F-score \(\geq0.45\); relative depth MAE \(\leq0.05\); normalized one-way Chamfer distance \(\leq0.03\); positive-depth ratio \(\geq0.98\); supported-surface ratio \(\geq0.50\) \\
\midrule
\rowcolor{TableGroup}[0pt][0pt]
\multicolumn{2}{@{}l@{}}{\textit{Memory and rendering}} \\
Surface anchors & Opacity \(\geq0.05\); 20-neighbor outlier removal with standard-deviation ratio 2.0; \(K_A=256\) anchors per object after FPS \\
Visibility composition & \(\tau_\alpha=0.05\), \(\eta=0.03\), and \(\epsilon=10^{-6}\) in Eqs.~\eqref{eq:appendix-relative-depth}--\eqref{eq:appendix-visibility} \\
\bottomrule
\end{tabularx}
\end{table*}

\noindent\textbf{World-space memory rendering.}
For every accepted object, the Sim(3) transform is applied to both the Gaussian
means and covariances:
\begin{equation}
\mu_{i,k}^{W}=s_iR_i\mu_{i,k}^{\mathrm{can}}+t_i,
\qquad
\Sigma_{i,k}^{W}=s_i^2R_i\Sigma_{i,k}^{\mathrm{can}}R_i^{\top}.
\label{eq:appendix-gaussian-world-transform}
\end{equation}

For completeness, let \(b_t^{\mathrm{obj}}(u)=\mathbf{1}[\alpha_t^{\mathrm{obj}}(u)>\tau_\alpha]\) and let \(b_t^{\mathrm{scene}}(u)\) denote the validity of the rendered background depth. When both are valid, their relative camera-depth difference is
\begin{equation}
\rho_t(u)
=
\frac{D_t^{\mathrm{obj}}(u)-D_t^{\mathrm{scene}}(u)}
{D_t^{\mathrm{scene}}(u)+\epsilon}.
\label{eq:appendix-relative-depth}
\end{equation}
Let
\(\mathbf{b}_t(u)=\bigl(b_t^{\mathrm{obj}}(u),b_t^{\mathrm{scene}}(u)\bigr)\) and
\(\mathbf{v}_t(u)=\bigl(v_t^{\mathrm{obj}}(u),v_t^{\mathrm{scene}}(u)\bigr)\).
We set the object and scene visibility gates to
\begin{equation}
\mathbf{v}_t(u)
=
\begin{cases}
(1,0), & \mathbf{b}_t(u)=(1,0),\\
(0,1), & \mathbf{b}_t(u)=(0,1),\\
(1,0), & \mathbf{b}_t(u)=(1,1),\ \rho_t<-\eta,\\
(0,1), & \mathbf{b}_t(u)=(1,1),\ \rho_t>\eta,\\
(0,0), & \text{otherwise}.
\end{cases}
\label{eq:appendix-visibility}
\end{equation}

\section{Implementation details}
\label{app:implementation-details}

\subsection{Memory construction}
\label{app:memory-parameters}

Following the memory construction pipeline in
Algorithm~\ref{alg:memory-construction}, we run VGGT-Omega-1B-512~\citep{wang2026vggtomega}
at its native input resolution of 512 pixels and map its pointmap back through
the recorded resize-and-crop transform before forming pixel correspondences.
Qwen3.5-122B-A10B~\citep{qwen2026qwen35} uses greedy decoding to return a single
structured list of the primary rigid
objects, including their names and normalized boxes. We exclude people, animals, fluids, and other evidently
non-rigid or deformable regions from the memory set. Each retained box is used
as the prompt to SAM3~\citep{meta2025sam3}, from which we retain the highest-confidence
associated instance mask. We run SAM3D~\citep{chen2025sam3d} with a fixed seed of
42. We retain at most eight selected objects per reference image. All remaining
construction settings, including the FPFH fallback~\citep{rusu2009fast} and validation thresholds, are
listed in Table~\ref{tab:memory-parameters}; distances are normalized by the
observed object scale \(d_i\).

\subsection{Model architecture}
\label{app:model-architecture}

\begin{table*}[t]
\centering
\caption{Architecture parameters.}
\label{tab:architecture-parameters}
\normalsize
\begin{tabularx}{\textwidth}{@{}>{\raggedright\arraybackslash}p{0.34\textwidth}H{19.0799pt}Y@{}}
\toprule
Setting & High-noise & Low-noise \\
\midrule
\rowcolor{TableGroup}[0pt][0pt]
\multicolumn{3}{@{}l@{}}{\textit{Expert-specific control}} \\
Control blocks & 20 & 10 \\
Paired expert blocks & \(\{2,4,\ldots,40\}\) & \(\{4,8,\ldots,40\}\) \\
Initialization & Copied from paired blocks & Copied from paired blocks \\
Object-centric refinement & Disabled & All 10 control blocks \\
\midrule
\rowcolor{TableGroup}[0pt][0pt]
\multicolumn{3}{@{}l@{}}{\textit{Shared condition encoding}} \\
Condition projections & \multicolumn{2}{>{\raggedright\arraybackslash\advance\leftskip by 8.8000pt\advance\rightskip by -8.8000pt}p{0.58\textwidth}@{}}{Separate \(3\times3\) projections \(P_{\mathrm{mem}}\) and \(P_{\mathrm{depth}}\); 16 output channels each} \\
Control-token embedding & \multicolumn{2}{>{\raggedright\arraybackslash\advance\leftskip by 8.8000pt\advance\rightskip by -8.8000pt}p{0.58\textwidth}@{}}{Stream concatenation followed by a \((1,2,2)\) patch embedding to dimension 5,120} \\
\midrule
\rowcolor{TableGroup}[0pt][0pt]
\multicolumn{3}{@{}l@{}}{\textit{Object-centric attention}} \\
Token assignment & \multicolumn{2}{>{\raggedright\arraybackslash\advance\leftskip by 8.8000pt\advance\rightskip by -8.8000pt}p{0.58\textwidth}@{}}{Dominant-object support \(\geq0.60\); valid canonical-coordinate coverage \(\geq0.50\)} \\
Anchor association & \multicolumn{2}{>{\raggedright\arraybackslash\advance\leftskip by 8.8000pt\advance\rightskip by -8.8000pt}p{0.58\textwidth}@{}}{Four nearest anchors; coordinates normalized by robust object radius; \(\tau_x=0.01\)} \\
Surface attention & \multicolumn{2}{>{\raggedright\arraybackslash\advance\leftskip by 8.8000pt\advance\rightskip by -8.8000pt}p{0.58\textwidth}@{}}{40 heads with head dimension 128} \\
Relation bias & \multicolumn{2}{>{\raggedright\arraybackslash\advance\leftskip by 8.8000pt\advance\rightskip by -8.8000pt}p{0.58\textwidth}@{}}{13D input; \(13\!\rightarrow\!128\!\rightarrow\!40\) SiLU MLP} \\
Relation scaling & \multicolumn{2}{>{\raggedright\arraybackslash\advance\leftskip by 8.8000pt\advance\rightskip by -8.8000pt}p{0.58\textwidth}@{}}{\(\lambda_{\mathrm{rel}}^{(l,h)}=1\) at initialization} \\
\bottomrule
\end{tabularx}
\end{table*}

\noindent\textbf{Backbone.}
We use the frozen two-expert Wan2.2-I2V-A14B backbone~\citep{wan2025wan,wanai2025wan22i2vdiffusers,huggingface2026diffusers}.
Each expert contains
40 transformer blocks with hidden dimension 5,120, 40 attention heads, and a
spatial patch size of \(2\times2\). The VAE uses a temporal-spatial stride of
\((4,8,8)\). Patchified Pl\"ucker-ray features~\citep{he2024cameractrl,bahmani2024ac3d}
are projected into token-aligned
camera embeddings. In each block, a
two-layer SiLU injector with a residual connection refines these embeddings,
and separate linear heads predict per-token scale and shift parameters that
modulate the hidden state after self-attention. The Pl\"ucker projection and
two-layer injectors use Xavier initialization, whereas the scale and shift
heads are zero-initialized, making the initial camera modulation an identity
mapping.

\noindent\textbf{Control branch.}
Before VAE encoding, scene depth is normalized once per video. Let
\(\ell_t(u)=\log(D_t^{\mathrm{scene}}(u)+\epsilon)\) at valid pixels, and let
\(q_2\) and \(q_{98}\) be the 2nd and 98th percentiles of \(\ell_t(u)\) over
all valid pixels in the rendered sequence. Define the clipped, normalized
log-depth as \(\hat{\ell}_t(u)=\bigl(\operatorname{clip}(\ell_t(u),q_2,q_{98})-q_2\bigr)/(q_{98}-q_2+\epsilon)\),
with \(\epsilon=10^{-6}\). We use
\begin{equation}
\bar{D}_t^{\mathrm{scene}}(u)=M_t^{\mathrm{scene}}(u)\left[2\hat{\ell}_t(u)-1\right].
\label{eq:appendix-depth-normalization}
\end{equation}
Thus, valid log-depth values are robustly mapped to \([-1,1]\), while invalid
pixels remain zero and are disambiguated by \(M_t^{\mathrm{scene}}\).

Using the fused control tokens \(H_c\) defined in
Eq.~\eqref{eq:condition-fusion}, we instantiate two expert-specific control
pathways with the configurations in Table~\ref{tab:architecture-parameters}.
Each control block reuses the timestep and text conditioning of its paired Wan
block. Within each pathway, the control state is propagated
sequentially, and every control-block output is connected through an independent
zero-initialized projection \(Z_{e,k}\) to its paired backbone state. For expert
\(e\in\{H,L\}\),
\begin{equation}
h_{b_{e,k}}^{\mathrm{Wan},e}
\leftarrow
h_{b_{e,k}}^{\mathrm{Wan},e}
+Z_{e,k}\!\left(h_{e,k}^{\mathrm{ctrl}}\right).
\label{eq:appendix-control-injection}
\end{equation}
Here, \(H\) and \(L\) denote the high- and low-noise experts, respectively,
and \(b_{e,k}\) is the \(k\)-th paired Wan block listed in
Table~\ref{tab:architecture-parameters}. At each denoising step, only one Wan
expert and its corresponding control pathway are evaluated.

In the control blocks enabled for object-centric refinement
(Table~\ref{tab:architecture-parameters}), canonical-space surface binding and
world-space relation preservation are inserted after the original self- and
cross-attention operations and before the feed-forward network. Both modules
preserve the hidden width and use independent zero-initialized
output projections, so their introduction does not perturb the pretrained
control computation at initialization. Tokens that do not satisfy both
assignment thresholds bypass these modules and remain on the original control
path.

\begin{table*}[t]
\centering
\caption{Memory construction statistics.}
\label{tab:memory-construction-statistics}
\normalsize
\begin{tabularx}{\linewidth}{@{}>{\raggedright\arraybackslash}p{0.38\linewidth}H{34.6700pt}H{10.3800pt}Y@{}}
\toprule
Statistic & RealEstate10K & DL3DV & Overall \\
\midrule
\rowcolor{TableGroup}[0pt][0pt]
\multicolumn{4}{@{}l@{}}{\textit{Dataset and memory counts}} \\
Training clips & 39,864 & 40,127 & 79,991 \\
Candidate objects & 197,247 & 215,322 & 412,569 \\
Objects per clip & 4.948 & 5.366 & 5.158 \\
Accepted memories & 184,067 & 193,874 & 377,941 \\
Valid memories per clip & 4.617 & 4.832 & 4.725 \\
Object success rate & 93.32\% & 90.04\% & 91.61\% \\
\midrule
\rowcolor{TableGroup}[0pt][0pt]
\multicolumn{4}{@{}l@{}}{\textit{Stage-wise construction outcomes}} \\
Reconstruction failures & 1.76\% & 2.32\% & 2.05\% \\
Initialization failures & 2.75\% & 4.19\% & 3.50\% \\
Validation failures & 2.03\% & 3.36\% & 2.72\% \\
Other failures & 0.14\% & 0.09\% & 0.12\% \\
\midrule
\rowcolor{TableGroup}[0pt][0pt]
\multicolumn{4}{@{}l@{}}{\textit{Fallback and coverage}} \\
FPFH triggered & 5.48\% & 7.03\% & 6.29\% \\
FPFH recovered & 65.74\% & 61.61\% & 63.34\% \\
Scene usable rate & 99.31\% & 98.62\% & 98.96\% \\
Target-view availability & 90.39\% & 86.15\% & 88.27\% \\
\bottomrule
\end{tabularx}
\end{table*}

\begin{table*}[t]
\centering
\caption{Extended ablation results.}
\label{tab:extended-ablation}
\footnotesize
\newcommand{\extendedmetrichead}[2]{\tablehead{#1\,#2}}
\newcommand{\extendedtwolinemetrichead}[3]{\tablehead{#1\\#2}\,#3}
\begin{equalspacedtabular}[0.1275pt]{6}{>{\raggedright\arraybackslash}m{0.30\textwidth}S{6.9572pt}cS{0.1415pt}cS{-0.0995pt}cS{0.1404pt}cS{-0.0960pt}cS{-0.0985pt}c}
\toprule
\tablehead{Variant} &
\extendedmetrichead{PSNR}{\raisebox{-.3pt}[0pt][0pt]{\(\uparrow\)}} &
\extendedmetrichead{SSIM}{\raisebox{-.3pt}[0pt][0pt]{\(\uparrow\)}} &
\extendedmetrichead{LPIPS}{\raisebox{-.3pt}[0pt][0pt]{\(\downarrow\)}} &
\extendedtwolinemetrichead{\kern-.06pt Object\kern.06pt}{Similarity}{\raisebox{-.15pt}[0pt][0pt]{\(\uparrow\)}} &
\raisebox{-.8pt}{\extendedtwolinemetrichead{\kern-.125pt Chamfer\kern.125pt}{Distance}{\raisebox{.68pt}[0pt][0pt]{\(\downarrow\)}}} &
\extendedmetrichead{MEt3R}{\raisebox{-.3pt}[0pt][0pt]{\(\downarrow\)}} \\
\midrule
\rowcolor{TableGroup}[0pt][0pt]
\multicolumn{7}{@{}l@{}}{\textit{View-aligned guidance}} \\
Wan2.2~\citep{wan2025wan,wanai2025wan22i2vdiffusers} Base & 15.86 & 0.473 & 0.521 & 47.2 & 0.312 & 0.386 \\
+ Camera Branch & 17.31 & 0.528 & 0.466 & 48.5 & 0.289 & 0.361 \\
+ Camera Branch + Projected Point Cloud & 21.04 & 0.653 & 0.347 & 53.7 & 0.234 & 0.284 \\
+ Camera Branch + Object-Centric Memory & 22.28 & 0.738 & 0.317 & 54.1 & 0.212 & 0.219 \\
+ Camera Branch + Dual-Stream Guidance & 23.15 & 0.759 & 0.241 & 58.6 & 0.173 & 0.208 \\
\midrule
\rowcolor{TableGroup}[0pt][0pt]
\multicolumn{7}{@{}l@{}}{\textit{Object-centric attention (added to Camera Branch + Dual-Stream Guidance)}} \\
+ Canonical-Space Surface Binding & 24.83 & 0.796 & 0.189 & 62.9 & 0.145 & 0.171 \\
+ World-Space Relation Preservation & 23.96 & 0.781 & 0.214 & 60.8 & 0.139 & 0.183 \\
\textbf{+ Object-Centric Attention} & \textbf{25.67} & \textbf{0.838} &
\textbf{0.139} & \textbf{65.3} & \textbf{0.124} & \textbf{0.138} \\
\bottomrule
\end{equalspacedtabular}
\end{table*}

\subsection{Training and inference}
\label{app:training-inference}

For neural network training, we use AdamW with
\(\beta_1=0.9\), \(\beta_2=0.95\), \(\epsilon=10^{-8}\), weight decay 0.01,
and a maximum gradient norm of 1.0. The network learning rate increases linearly
to \(10^{-5}\) over the first 500 iterations and remains constant thereafter.
The short per-object geometric refinement uses Adam with the settings listed in
Table~\ref{tab:memory-parameters}.
Each GPU processes one 81-frame clip, and gradients are accumulated for four
steps across eight A800 GPUs, giving an effective batch size of 32. We use bf16
precision and the original Wan2.2 flow-matching objective without additional
reconstruction losses.

During the first 8,000 iterations, both Wan experts, the VAE, and the text
encoder are frozen, while the camera branch, \(P_{\mathrm{mem}}\),
\(P_{\mathrm{depth}}\), \(f_{\mathrm{emb}}^c\), and the 30 expert-specific
control blocks are optimized. For the following 5,000 iterations, we freeze
the camera branch, continue optimizing the two control pathways and their
condition encoders, and enable object-centric attention only in the low-noise
pathway. We independently drop the text condition with probability 0.1 and jointly
drop the object-memory and scene-depth streams with probability 0.1, while
always retaining the camera condition.

At inference, we use FlowUniPC with 50 denoising steps, a flow shift of 5.0,
and classifier-free guidance scales of 3.5 for both the high- and low-noise
experts. Camera and rendered-memory condition scales are both 1.0. Unless
otherwise specified, Totem generates 81 frames at 15 fps and 720p resolution.

\begin{figure}[H]
    \centering
    \includegraphics[width=\columnwidth]{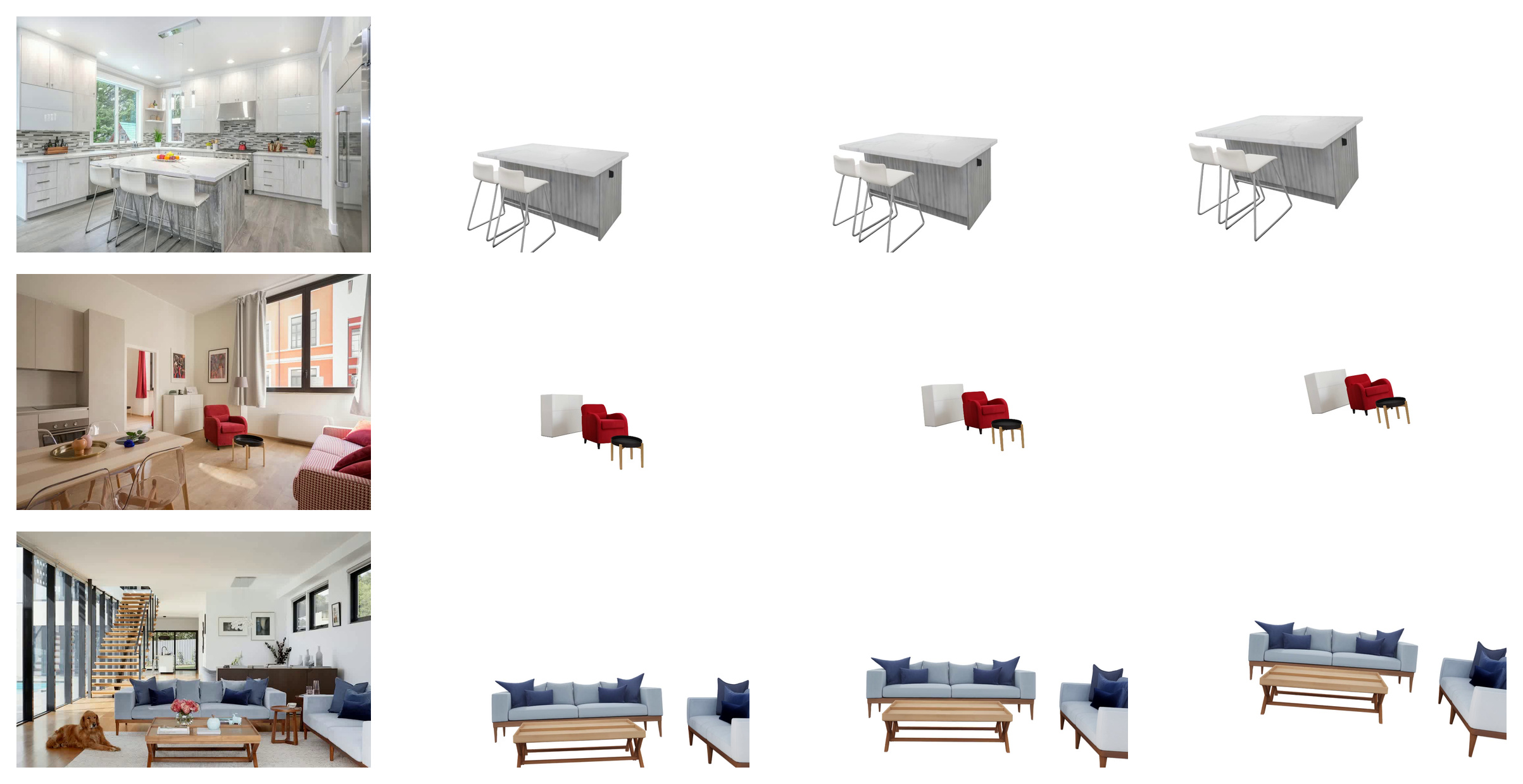}
    \par\vspace{\floatcontentsep}
    \paperfigurelabel{\scriptsize%
    \noindent\hbox to \linewidth{%
    \hbox to 0.25\linewidth{\hfil Reference frame\hfil}%
    \hbox to 0.25\linewidth{\hfil Full alignment\hfil}%
    \hbox to 0.25\linewidth{\hfil Initialization only\hfil}%
    \hbox to 0.25\linewidth{\hfil No alignment\hfil}%
    }}
    \caption{Qualitative 3D memory alignment results. From left to right: the reference frame, fully aligned memory, initialization-only memory, and unaligned memory.}
    \label{fig:memory-alignment-qualitative-figure}
\end{figure}

\begin{figure*}[t]
    \centering
    \setlength{\abovecaptionskip}{\floatcontentsep}
    \includegraphics[trim=0 17.5pt 0 0,clip,width=\linewidth]{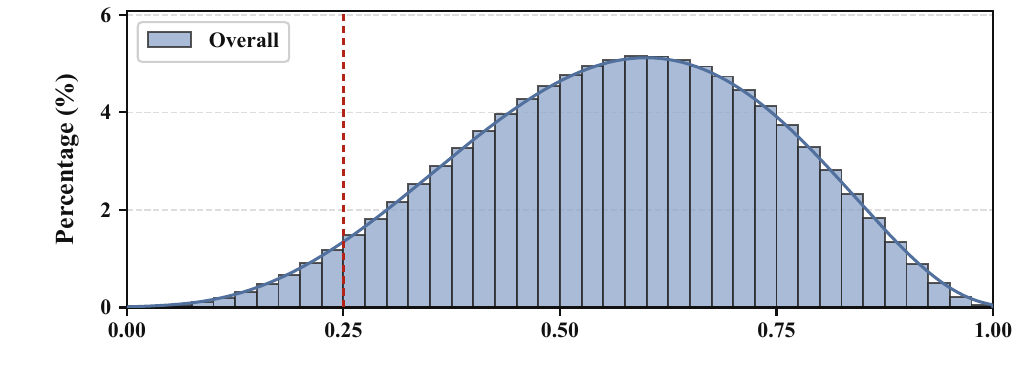}
    \par\vspace{\floatcontentsep}
    {\paperfigurelabel{(a) RANSAC}\par}
    \vspace{\floatcontentsep}
    \includegraphics[trim=0 17.5pt 0 0,clip,width=\linewidth]{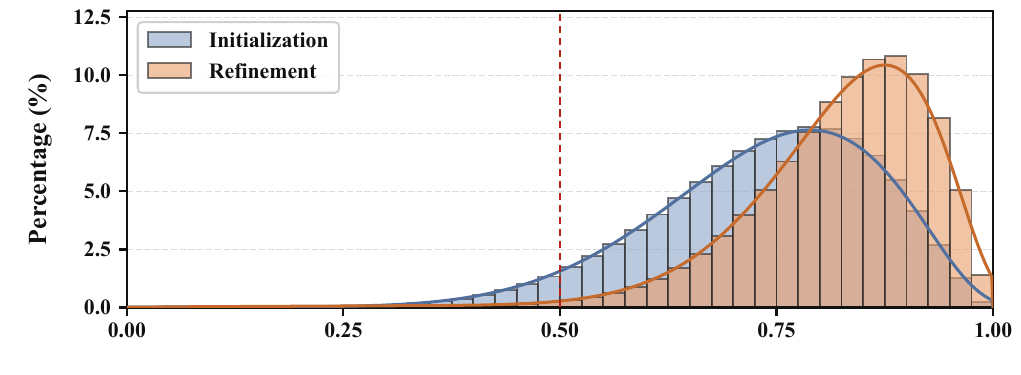}
    \par\vspace{\floatcontentsep}
    {\paperfigurelabel{(b) Mask IoU}\par}
    \vspace{\floatcontentsep}
    \includegraphics[trim=0 17.5pt 0 0,clip,width=\linewidth]{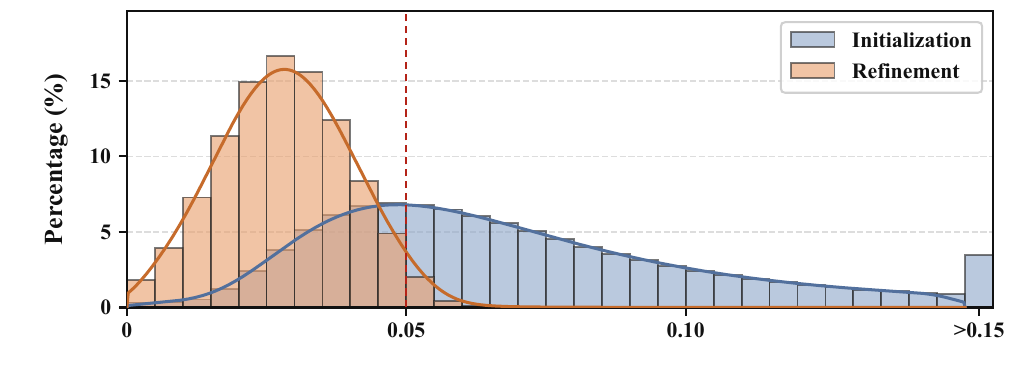}
    \par\vspace{\floatcontentsep}
    {\paperfigurelabel{(c) Depth MAE}\par}
\caption{Memory construction quality distributions.}
\label{fig:memory-construction-distributions}
\end{figure*}

\begin{figure*}[t]
    \centering
    \includegraphics[width=\textwidth]{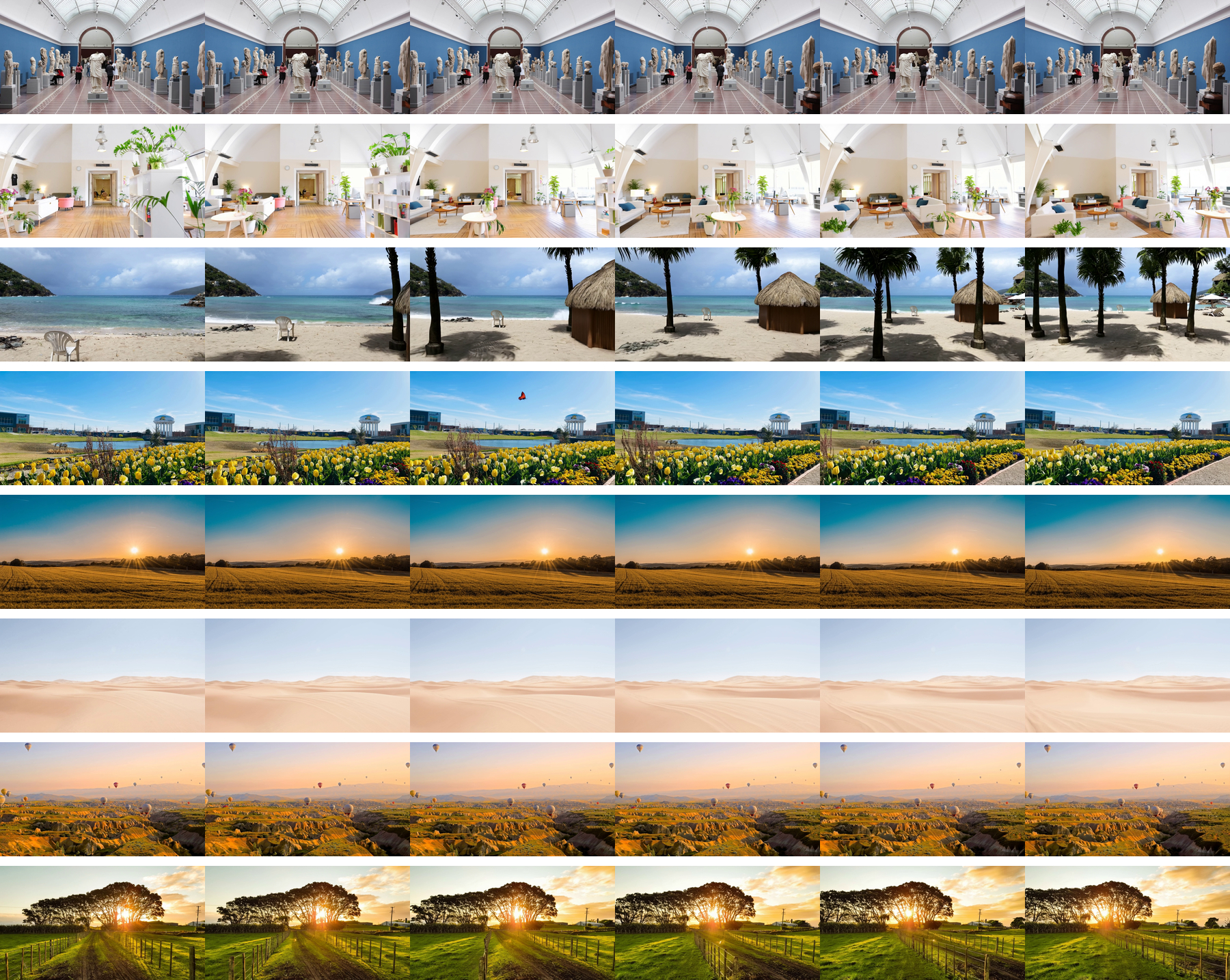}
    \caption{Additional qualitative results.}
    \label{fig:additional-results-grid}
\end{figure*}

\section{Analysis of memory construction}
\label{app:memory-analysis}

Memory construction is evaluated on all 79,991 training clips from
RealEstate10K~\citep{zhou2018stereo} and DL3DV~\citep{ling2024dl3dv}.
Table~\ref{tab:memory-construction-statistics}
summarizes dataset-level counts, stage-wise outcomes, and coverage
statistics; percentages are computed over candidate objects, whereas
accepted-memory entries are reported as absolute counts. Figure~
\ref{fig:memory-construction-distributions} shows pooled alignment-quality
distributions. Panel (a) uses defined pixel-correspondence RANSAC inlier
ratios only, and FPFH recoveries do not add RANSAC observations. Panels (b)
and (c) compare initialization and refinement on the same 389,671 alignments
that pass initialization, including successful FPFH recoveries.

The median RANSAC inlier ratio is 0.58 (IQR 0.45--0.70). Local refinement
increases the median reference-view Mask IoU from 0.75 to 0.84 and reduces the
median relative depth MAE from 0.063 to 0.028.

Scene usability denotes the fraction of clips containing at least one accepted
memory, while target-view availability denotes the fraction of target frames
with a valid visibility-composed memory mask. Records without a stage-specific
diagnostic are categorized as ``other failures.''

\section{Additional results}
\label{app:additional-results}

\subsection{Qualitative results}
\label{app:additional-qualitative-results}

Figure~\ref{fig:additional-results-grid} presents representative successful
clips selected from the application set. Each row corresponds to one
video, and the six columns show evenly spaced frames from the 81-frame clip;
the rows cover indoor, coastal, suburban, and open-landscape scenes.

\subsection{Quantitative results}
\label{app:additional-quantitative-results}

To complement the ablation study in the main paper, Table~\ref{tab:extended-ablation}
provides a finer-grained analysis of the conditioning design on the OOD benchmark.
We compare camera control, projected point-cloud guidance, and Dual-Stream
guidance as individual and combined conditions, and then evaluate Canonical-Space
Surface Binding and World-Space Relation Preservation individually and jointly.

\section{Limitations and future work}
\label{app:limitations}

Our current memory is reconstructed as a static 3D representation from the
reference view. It is therefore best suited to scenes whose geometry and object
appearance remain approximately rigid during the generated clip. In dynamic
scenes, independently moving objects, articulated motion, and non-rigid changes
can make the fixed object-to-world placement and canonical surface addresses
stale, reducing temporal consistency and producing incorrect occlusions or
textures. Extending the memory from a static 3D representation to a time-aware
4D representation is a natural next step for handling such changes.